%% file: main.tex
\documentclass{article}

\usepackage{natbib}
\usepackage{arxiv}

\usepackage[utf8]{inputenc} 
\usepackage[T1]{fontenc}    
\usepackage{url}            
\usepackage{booktabs}       
\usepackage{amsfonts}       
\usepackage{nicefrac}       
\usepackage{microtype}      
\usepackage{graphicx}

\usepackage{hyperref}       
\usepackage{doi}
\usepackage{amsmath}
\usepackage{amssymb}
\usepackage{algorithm}
\usepackage{algorithmic}
\usepackage{rotating, booktabs}
\usepackage{tikz, pgfplots}
\pgfplotsset{compat=1.18}
\usepackage{tabularray}
\usepackage{array}
\usepackage{multirow}
\usepackage{hyperref}

\usepackage{graphicx}
\usepackage{booktabs}
\usepackage{multirow}
\usepackage{array}
\usepackage{pifont}
\usepackage{placeins}

\title{What really matters for person re-identification? A Mixture-of-Experts Framework for Semantic Attribute Importance}

\date{} 					

\author{ \href{https://orcid.org/0000-0002-7412-0529}{\includegraphics[scale=0.06]{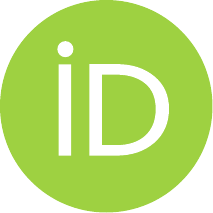}\hspace{1mm}Athena Psalta} \\
	Remote Sensing Laboratory\\
	National Technical University of Athens\\
	Iroon Polytechneiou 9, Athens 15780, Greece \\
	\texttt{psaltaath@central.ntua.gr} \\
	\And
	\href{https://orcid.org/0000-0003-2592-2127}{\includegraphics[scale=0.06]{orcid.pdf}\hspace{1mm}Vasileios Tsironis} \\
	Remote Sensing Laboratory\\
	National Technical University of Athens\\
	Iroon Polytechneiou 9, Athens 15780, Greece \\
	\texttt{tsironisbi@central.ntua.gr} \\
    \And
	\href{https://orcid.org/0000-0001-8730-6245}{\includegraphics[scale=0.06]{orcid.pdf}\hspace{1mm}Konstantinos Karantzalos} \\
	Remote Sensing Laboratory\\
	National Technical University of Athens\\
	Iroon Polytechneiou 9, Athens 15780, Greece \\
	\texttt{karank@central.ntua.gr} \\
}

\renewcommand{\headeright}{Preprint}
\renewcommand{\undertitle}{Preprint}
\renewcommand{\shorttitle}{MoSAIC-ReID}

\hypersetup{
pdftitle={A template for the arxiv style},
pdfsubject={q-bio.NC, q-bio.QM},
pdfauthor={David S.~Hippocampus, Elias D.~Striatum},
pdfkeywords={First keyword, Second keyword, More},
}

\begin{document}
\maketitle

\begin{abstract}
State-of-the-art person re-identification methods achieve impressive accuracy but remain largely opaque, leaving open the question: which high-level semantic attributes do these models actually rely on? We propose MoSAIC-ReID, a Mixture-of-Experts framework that systematically quantifies the importance of pedestrian attributes for re-identification. Our approach uses LoRA-based experts, each linked to a single attribute, and an oracle router that enables controlled attribution analysis. While MoSAIC-ReID achieves competitive performance on Market-1501 and DukeMTMC under the assumption that attribute annotations are available at test time, its primary value lies in providing a large-scale, quantitative study of attribute importance across intrinsic and extrinsic cues. Using generalized linear models, statistical tests, and feature-importance analyses, we reveal which attributes, such as clothing colors and intrinsic characteristics, contribute most strongly, while infrequent cues (e.g. accessories) have limited effect. This work offers a principled framework for interpretable ReID and highlights the requirements for integrating explicit semantic knowledge in practice. Code is available at \url{https://github.com/psaltaath/MoSAIC-ReID}

\end{abstract}

\keywords{Person Re-Identification \and Mixture of Experts \and Low-Rank Adaptation \and Attribute Importance}

\input{sec/introduction}

\input{sec/relatedwork}
\input{sec/methodology}
\input{sec/experimentalresults}
\input{sec/conclusions}
\input{sec/appendix}

\bibliographystyle{bibstyles}
\bibliography{main.bib}

\end{document}

%% file: sec/introduction.tex
\section{Introduction}
\label{introduction}
Person re-identification is a fundamental computer vision task related to critical applications such as transportation analytics (\cite{yu2023semantic, behera2023large}) and multiple object tracking (\cite{psalta2024transformer, du2023strongsort}). Its objective is to reliably match pedestrian images across non-overlapping camera views, despite variations in appearance, lighting and occlusion. Recent approaches (\cite{che2025enhancing, chen2023beyond, somers2024keypoint}) have set new standards for re-identification accuracy and generalization by leveraging vision-language models and auxiliary semantic information, such as textual descriptions or pedestrian attributes. However, as these models have grown in complexity and capacity, a persistent challenge remains: \textit{Which semantic attributes actually contribute to successful person retrieval, and to what extent?} 

Most re-identification methods provide little insight into which pedestrian attributes drive identification decisions or how robust the system is to changes in these attributes. Classic post-hoc interpretability tools or global feature importance analyses are limited, often conflating correlated cues and failing to establish direct, causal relationships between semantic attributes and model outcomes. This motivates the need for a systematic, modular approach that can disentangle and quantify the role of individual attributes in identification performance.

\begin{figure*}[t]
    \centering
    \includegraphics[width=\textwidth]{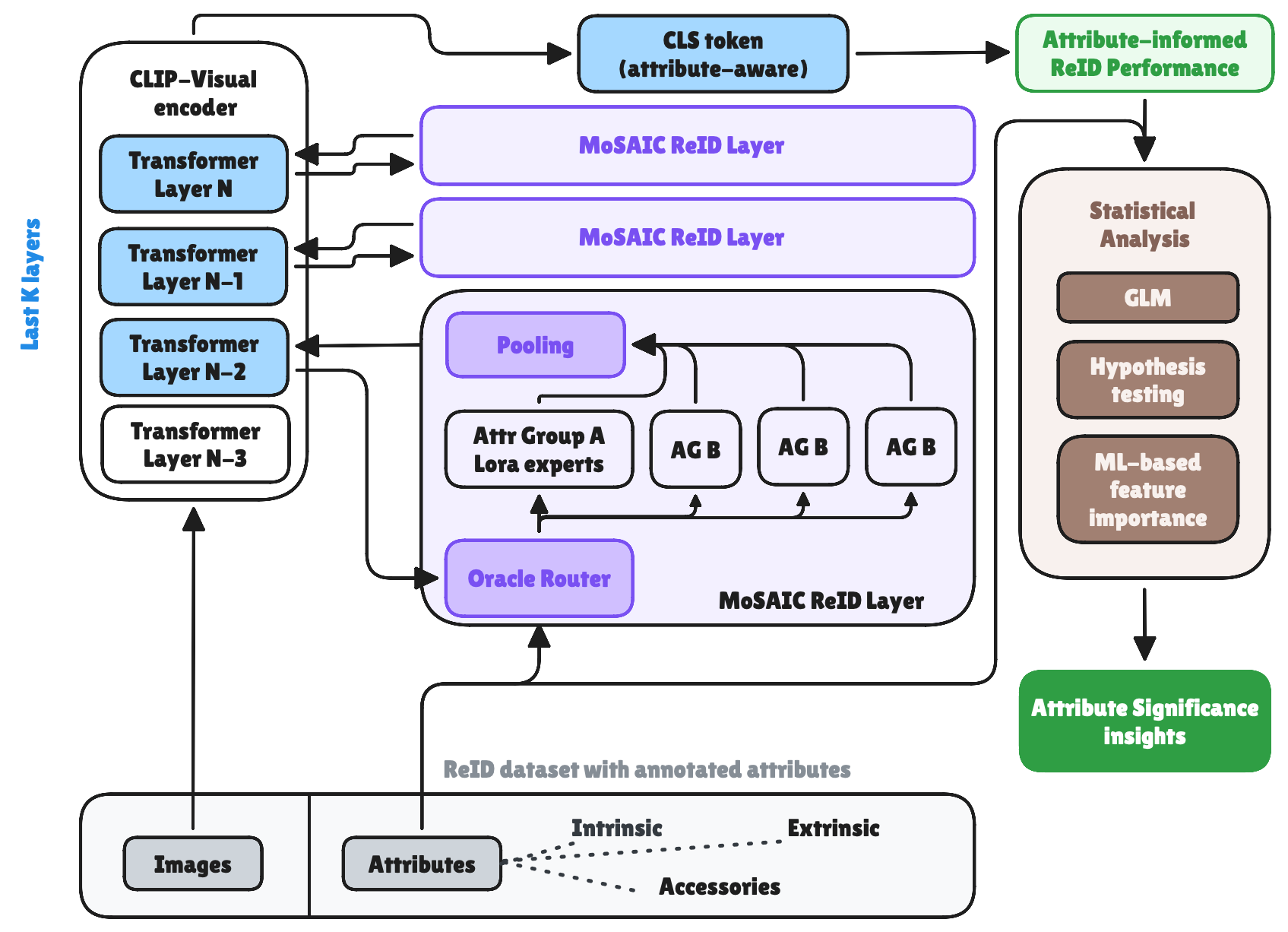}
    \caption{\textbf{MoSAIC‑ReID framework}. A CLIP‑based visual encoder is augmented in its last transformer layers with MoSAIC‑ReID modules composed of semantic LoRA experts grouped by attribute type and activated through an oracle router using ground‑truth annotations. The resulting attribute‑aware CLS token yields attribute‑informed ReID performance, which is subsequently analysed through generalized linear models, hypothesis testing and ML‑based feature importance to derive quantitative insights on semantic attribute significance for re-identification.}
    \label{fig:overview}
\end{figure*}

In this work, we propose Mixture of Semantic Attribute-Informed Components for Re-Identification (MoSAIC-ReID), a novel Mixture-of-Experts (MoE) (\cite{jordan1994hierarchical}) framework to address these challenges by explicitly decomposing visual semantic space within transformer-based models for person re-identification (Figure \ref{fig:overview}). Our approach structures each expert as a LoRA (\cite{lora}) module, specialized for a single semantic attribute, and introduces a deterministic oracle routing mechanism. This allows for precise, ablation-style measurement; experts can be selectively activated or deactivated, enabling isolated assessment of marginal attribute contributions to accuracy. Unlike conventional post-hoc strategies, our method supports direct, causal inference about the value of  specific attribute knowledge at inference time. MoSAIC-ReID is architecturally flexible, integrating seamlessly with any Transformer-based re-identification backbone and supporting a wide range of semantic attributes, data domains, and evaluation scenarios. As a result, our conclusions about attribute significance are robust, reproducible, and not tied to any particular model implementation. Specifically, our contributions are: 

\begin{itemize}
\item We introduce a modular, attribute-specialized MoE framework for interpretable analysis of semantic cues for transformer-based person re-identification methods.
\item Our approach enables principled statistical analysis of attribute importance, quantifying the isolated impact of each semantic attribute through expert-wise ablation and oracle routing.
\item We perform a comprehensive evaluation on challenging benchmarks with detailed attribute annotations, combining GLMs, ML-based feature importance, and statistical hypothesis testing.
\item Our results yield new insights into which semantic attributes matter most for identity discrimination, informing the future design and deployment of more transparent, robust, and adaptive re-identification systems.
\end{itemize}

It is important to note that our contribution is methodological; we introduce MoSAIC-ReID as a framework to quantify and interpret attribute importance in ReID, not as a deployable real-time system, with an oracle-based router that highlights a principled way to isolate attributes’ causal effects.

%% file: sec/relatedwork.tex
\section{Related Work}
\label{relatedwork}

\textbf{Person re-identification.}  Recent advances in person re-identification have shown that Transformer-based (\cite{vaswani2017attention}) and vision-language models such as CLIP (\cite{radford2021learning}) have set new performance standards in both accuracy and versatility. CLIP-based approaches have achieved strong cross-modal feature alignment even without explicit textual labels (\cite{li2023clip, yan2023clip}). Recent works extend these ideas by exploiting text inversion mechanisms (\cite{yang2024pedestrian, wang2025idea}), unifying pedestrian attribute representation through explicit and implicit text prompts (\cite{zhai2024multi}) or combining CLIP with sequential modeling and aggregation to bridge image-language domains within a hybrid architecture (\cite{yu2025climb}). Since pedestrian attributes usually stand as robust visual properties in condition changes, early multi-task learning approaches (\cite{market-attr, dukemtmc-attr}) established that joint optimization of identity and attribute prediction improves the discriminative power of learned representations by exposing the model to both global identity and fine-grained attribute information. Similarly, subsequent approaches (\cite{huang2024attribute, ahmed2025multi, eom2025cerberus}) have showcased that incorporating semantic attribute cues enhances retrieval accuracy. With the advancement of vision-language models, access to reliable and explicit attribute information at inference time is becoming increasingly feasible, enabling more structured and interpretable semantic analysis. However, most existing attribute-based methods rely on post-hoc analysis or black-box feature importance techniques, which often lack interpretability and cannot provide direct, causal insights into the role of individual attributes. In contrast, our framework enables a principled statistical analysis that rigorously quantifies the significance of each semantic human attribute.

\noindent\textbf{Mixture of Experts (MoE)} (\cite{jordan1994hierarchical}) has emerged as a powerful architectural paradigm across domains, such as computer vision (\cite{Chen_2025_CVPR, Cai_2025_CVPR, Rahman_2025_WACV}), large language models (\cite{zhu2024llama, li2024cumo}) and multimodal processing (\cite{Liu_2025_CVPR, wu2024deepseek}). At its core, MoE addresses the complexity of learning tasks through a dynamic or learned routing mechanism and allocates input data or sub-tasks to distinct expert modules for capacity expansion and representational diversity without proportionally increasing computational overhead. Recent developments demonstrate that sparsely activated experts —either through static routing (\cite{fedus2022switch}) or adaptive policies (\cite{zhou2022mixture, dai2024deepseekmoe, yun2024flex})— can greatly enhance scalability and robustness. Also, the combination of Parameter-Efficient Fine-Tuning (PEFT) techniques like Low-Rank Adaptation (LoRA) (\cite{lora}) with MoE has proven especially effective for large models, enabling dynamic specialization across layers (\cite{gao2024higher}) and improved multi-task generalization through task-aware routing and clustering strategies (\cite{wu2024mixture}). Contemporary re-identification methods leverage expert selection or multi-modal feature specialists within the re-identification pipeline by dynamically routing features to appropriate experts (\cite{dai2021generalizable}), enabling specialized clusters of experts to capture specific data distributions (\cite{ren2025mosce}) or separating appearance and modality cues through a decoupled MoE framework (\cite{wang2025decoupled}). Building on insights from these advancements, our approach adapts MoE paradigm into a Transformer-based backbone to address the specific challenges of interpretability and causal attribution in person re-identification. Rather than relying on generic or task-blind expert modules, each expert purposefully focuses on distinct semantic attributes for explicit modeling of attribute contributions and their interactions, while the proposed routing mechanism allows precise control and ablation to facilitate causal analysis.

%% file: sec/methodology.tex
\section{Methodology}
\label{methodology}

\subsection{Prerequisites}
\label{methodology_prerequisites}

\noindent\textbf{Problem Formulation.} Re-identification aims to retrieve images of a target individual captured by different, non-overlapping views. Formally, given a query image $q$, the goal is to identify images of the same identity from a gallery set $\mathcal{G} = \{g_1, g_2, \ldots, g_N\}$, where the person depicted in $q$ may appear under different viewpoints and illumination conditions. Each image $x \in \{q\} \cup \mathcal{G}$ is encoded into a visual embedding vector $f(x) \in \mathbb{R}^d$ by a backbone encoder. Retrieval is performed by ranking gallery embeddings based on their similarity to the query embedding. Standard evaluation metrics include mean average precision (mAP) and Cumulative Matching Characteristics (CMC) at rank-1 accuracy, which quantify both retrieval quality and ranking correctness.

\noindent\textbf{CLIP-ReID (\cite{li2023clip})} adapts the CLIP model for re-identification tasks lacking explicit text annotations. CLIP-ReID introduces a two-stage training strategy to leverage the cross-modal capabilities of CLIP. In the first stage, learnable ID-specific text tokens are introduced into the template \textit{``A photo of a $S^\ast$ person''}, where $S^\ast$ denotes a sequence of $M$ learnable tokens initialized randomly. Only these tokens are optimized using contrastive image-text losses, while the image and text encoders remain frozen. In the second stage, the optimized text tokens are held fixed and used to regularize the fine-tuning of the image encoder through a combination of identity classification loss, triplet loss and a text-guided image-to-text cross-entropy loss for aligning image embeddings with learned semantic concepts in a discriminative manner.

\begin{figure*}[t]
    \centering
    \includegraphics[width=0.8\textwidth]{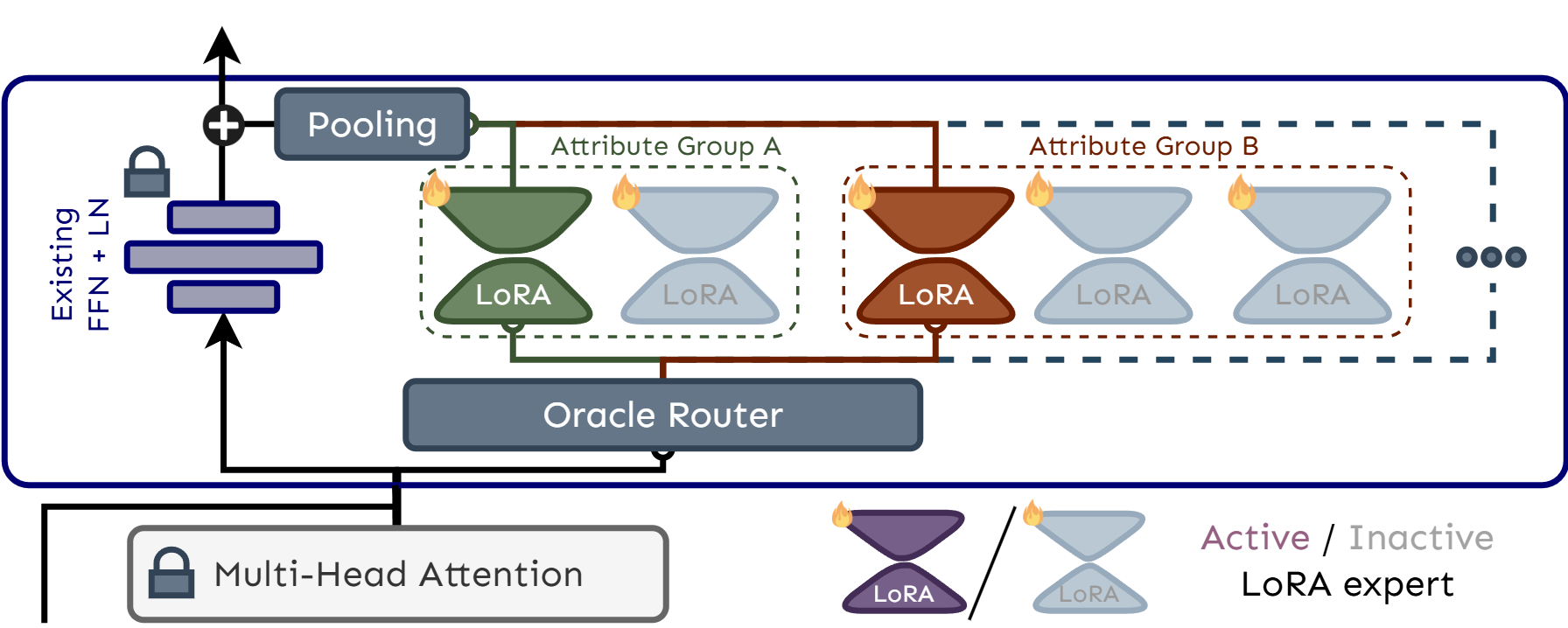}
    \caption{\textbf{Overview of the MoSAIC-ReID architecture} which can be integrated within a transformer-based visual encoder. LoRA experts are organized into semantic groups, each aligned with a specific attribute. An oracle router deterministically activates experts based on ground-truth attributes, enabling explicit attribute-aware representation learning. Expert outputs are aggregated with a pooling mechanism and combined via a residual connection, ensuring both the original and attribute-enhanced features contribute to the final embedding.}
    \label{fig:moe-arch}
\end{figure*}

\subsection{Overview and place-in-architecture}
\label{methodology_overview}

Our approach introduces a novel MoE module as a core component within the visual encoder of a transformer-based architecture. Specifically, we integrate our module into the visual transformer encoder of CLIP (\cite{radford2021learning}), targeting the last $K$ layers. In these layers, the standard feed-forward network (FFN) is overridden by our residual MoE design, allowing for attribute-aware representation while preserving the original model’s capacity.

The MoE module is implemented in a residual configuration (Figure \ref{fig:moe-arch}). The original Feed-Forward Network (FFN) in the Transformer architecture is retained operating as usual, while the outputs of the newly introduced expert groups are injected via a skip connection. This ensures that the original information flow is preserved and the contribution of the experts is additive rather than substitutive for stable optimization and effective knowledge integration. 

\subsection{Semantic Expert Structure and Oracle Routing}
\label{methodology_routing}

MoSAIC-ReID introduces semantic expert groups and leverage a deterministic oracle router for activation for supporting interpretable attribute modeling. To capture fine-grained attribute information, we cluster our experts into semantic groups, each mapped to a single visual attribute. Thus, \textit{each group contains as many experts as attributes}. Attributes are categorized as follows:
\begin{itemize}
    \item \textbf{Single-state binary attributes:} Each group contains a single expert, suitable for attributes that may or may not be present (e.g., ``a person carries a bag''). If the attribute is absent, no expert is activated for that group.
    \item \textbf{Dual-state binary attributes:} Groups contain two experts, each representing one of the two possible states of the attribute (e.g., ``short sleeves'' vs. ``long sleeves''). For every sample, one expert is activated according to the observed state.
    \item \textbf{Multiclass attributes:} Groups contain three or more experts, each corresponding to a distinct category of the attribute (e.g., ``top color'' with experts for red, blue, green, etc.). The expert matching the sample’s value is activated. 
\end{itemize}

\begin{figure*}[t]
    \centering
    \includegraphics[width=0.8\textwidth]{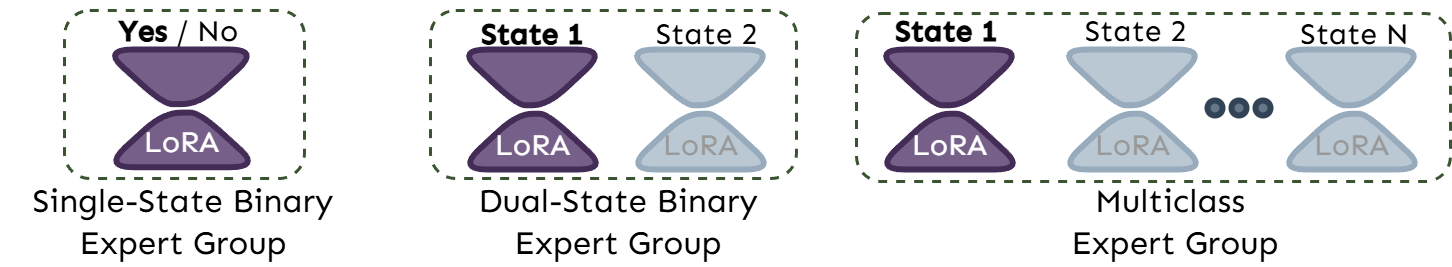}
    \caption{\textbf{Expert group activation for different attribute types.} Left: For single-state binary attributes, a single LoRA expert is activated only if the attribute exists. Middle: For dual-state binary attributes, one of two experts is activated based on the observed attribute state. Right: For multiclass attributes, exactly one expert in a group is activated according to the specific attribute category.}
    \label{fig:expert-states}
\end{figure*}

During the forward pass, expert activation is governed by an attribute-aware \textbf{oracle router}. The router has access to ground-truth attribute annotations for each input instance, both during training and inference, and deterministically selects the appropriate expert(s) within each semantic group. For single-state binary attributes (e.g., "wearing a hat"), the expert is only activated if the attribute is present; if not, the group remains inactive. For dual-state binary attributes (e.g., "short/long sleeves") and for multiclass attributes (e.g., age), exactly one expert is activated per group, corresponding to the specific state/category observed (Figure \ref{fig:expert-states}). Through this deterministic routing we ensure that the representation is explicitly conditioned on known semantic attributes to enable precise attribute-aware ReID. 

\subsection{Expert Parameterization and Aggregation}
\label{methodology_parameterization}

Each expert within a semantic group is implemented using a LoRA module (\cite{lora}), which allows parameter-efficient fine-tuning while significantly reducing the computational overhead commonly associated with large-scale MoE architectures. When an expert is activated by the oracle router, it processes the token sequence locally and outputs a single token that encodes attribute-specific information. In  contrast, if no expert is activated for a group (e.g., the associated attribute is absent), that group does not contribute any token to the final output. All tokens generated by the active experts across semantic groups are then aggregated using a pooling operation, such as mean or max pooling, to produce a single summary token. This token is subsequently added to the original token sequence after the FFN via a residual connection. In this way, MoSAIC-ReID ensures that both the base representation from the underlying transformer and the attribute-enhanced features from the experts are preserved in a unified form. Our design enables flexible and efficient incorporation of attribute-specific signals while maintaining full compatibility with pre-trained transformer structures, supporting modularity, interpretability and scalability across diverse attribute configurations.

\subsection{Training Strategy}
\label{methodology_training}

Our training strategy closely follows Stage 2 of the CLIP-ReID framework, with modifications designed to accommodate our attribute-aware MoE model. The key objective is to optimize only the LoRA-based expert modules within the visual encoder, while keeping both the original feed-forward networks and all other parameters of the CLIP visual transformer entirely frozen to allow the training process to focus sharply on infusing attribute-specific capabilities into the model. Throughout training of MoSAIC-ReID, we leverage the set of learnable textual prompts (the \textit{prompt learner}) obtained from Stage 1 of CLIP-ReID training. These prompts, which remain fixed during Stage 2, provide robust alignment between visual and textual modalities, as they supply enriched text embeddings specific to each identity.

During optimization, only the parameters of the LoRA experts are updated. All base transformer parameters and the prompt tokens remain frozen. The training process uses the fixed ID-specific prompts derived from Stage 1, harnessing their rich semantic content for text-image alignment. The core learning objective maximizes the similarity between visual and text features for matching identity pairs, while minimizing it for non-matching ones. Each image uses its associated, prompt-augmented text embedding as a positive reference, forming a contrastive learning setup.

Loss formulation in this stage mirrors the objectives of CLIP-ReID Stage 2, integrating three key components. The first is an identity classification loss with label smoothing ($\mathcal{L}_{\mathrm{ID}}$), calculated as a standard cross-entropy loss over smoothed identity probabilities:

\begin{equation}
    \mathcal{L}_{\mathrm{ID}} = - \sum_{k=1}^{N} q_k \log(p_k)
\end{equation}

\noindent where $q_k = (1-\epsilon)\cdot \mathbb{I}[k = y] + \frac{\epsilon}{N}$ is the smoothed ground truth distribution and $p_k$ represents the predicted identity probabilities.

To further enforce discriminative learning across identities, we apply a triplet loss ($\mathcal{L}_{\mathrm{tri}}$), which encourages margin-based separation between matched (positive) and unmatched (negative) image embeddings:

\begin{equation}
    \mathcal{L}_{\mathrm{tri}} = \max( d_p - d_n + m, 0 )
\end{equation}

\noindent where $d_p$ and $d_n$ are the feature distances between the positive and negative pairs, and $m$ is a margin hyperparameter.

An additional image-to-text contrastive loss with label smoothing ($\mathcal{L}_{\mathrm{i2t}}$) is employed to encourage alignment between the visual representation and its corresponding textual representation:

\begin{equation}
    \mathcal{L}_{\mathrm{i2t}}(i) = - \sum_{k=1}^{N} q_k \log \left(
        \frac{ \exp\left(s(\mathbf{v}_i, \mathbf{t}_k) \right)}{\sum_{a=1}^{N} \exp\left(s(\mathbf{v}_i, \mathbf{t}_a)\right)}
    \right)
\end{equation}

\noindent where $s(\cdot,\cdot)$ represents a similarity function (e.g., cosine similarity), and $\mathbf{v}_i$, $\mathbf{t}_k$ denote the image and text embeddings, respectively, with $q_k$ applying label smoothing.

The final training loss combines these three components:

\begin{equation}
    \mathcal{L}_{\mathrm{stage2}} = \mathcal{L}_{\mathrm{ID}} + \mathcal{L}_{\mathrm{tri}} + \mathcal{L}_{\mathrm{i2t}}
\end{equation}

Overall, this training regime preserves the full representational strengths and compositionality of CLIP while enabling efficient specialization to known semantic attributes for person re-identification. By maintaining a pure Stage 2 fine-tuning process —without any updates to the base CLIP encoder or prompt learner components— we retain CLIP's strong generalization ability while injecting the attribute-awareness that MoSAIC-ReID leverages for interpretable, modular and semantically grounded person re-identification. Notably, this training strategy does not depend on any architectural specifics of CLIP-ReID and can be seamlessly leveraged with any transformer-based visual encoder, enabling broad adaptability of MoSAIC-ReID to diverse person re-identification backbones.

%% file: sec/experimentalresults.tex
\section{Experimental Results}
\label{results}

\subsection{Datasets \& experimental protocol }
\label{results_datasets}

\noindent To investigate the influence of semantic attributes on person re-identification, we conducted experiments using two widely adopted benchmarks. 

\noindent \textbf{Market-1501} (\cite{market}) consists of over 32,000 bounding box images of 1,501 identities, captured across six cameras with realistic viewpoint and appearance variability, where each individual is represented, on average, by approximately 3.6 images per camera view. The dataset is divided into two subsets: 750 identities for the training phase and the remaining 751 identities for testing. Following the standard evaluation protocol, 3,368 images serve as query examples to retrieve correct matches from 19,732 gallery images. In this work, we utilize the expanded set of 27 hand-annotated attributes described in the work of Zhang et al. (\cite{market-attr}), which include gender, hair length, sleeve and clothing types, age, presence of accessories (bag, backpack, hat, backpack), and detailed color annotations for both upper and lower body garments. 

\noindent \textbf{DukeMTMC} (\cite{dukemtmc}) contains more than 36,000 manually cropped images of 1,404 identities. Sourced from high-resolution videos captured by eight cameras, DukeMTMC provides a challenging test bed with substantial variation in viewpoint, illumination, and background. The dataset includes a training split of 16,522 images for 702 identities, while another disjoint set of 702 identities forms the test partition, with 2,228 query images and 17,661 gallery images for evaluation. For this work, we rely on the comprehensive set of 23 hand-annotated attributes introduced in Lin et al. (\cite{dukemtmc-attr}), which expand the standard annotations to include properties like age, hair length, sleeve type, clothing length, footwear, bag and backpack presence, and several color categories for both upper and lower garments. 

\noindent \textbf{Our dataset choice reflects availability}: Market-1501 and DukeMTMC are the only benchmarks with usable attribute annotations, but we note the importance of extending to newer datasets as annotations become available. We also acknowledge broader concerns such as fairness of demographic attributes, however our efforts are restricted and limited to the demographics of these two datasets. For example, IUST dataset (\cite{IUST}) is a major effort to reduce demographic bias in the ReID task, however there are not attribute-level annotations available for that or similar-scope datasets, thus cannot be included in this analysis.
\input{tables/attributes.tex}

To systematically assess the impact of attribute information at inference, we designed a series of experiments for both datasets. The experiments involved training and evaluating MoSAIC-ReID with various combinations of attribute groups either included or excluded. Attributes were clustered into five semantically coherent categories, presented in Table \ref{table-attr}: \textbf{Intrinsic} (e.g., gender, age), \textbf{Upper-body (Up)} (e.g., sleeve length, color), \textbf{Lower-body (Down)} (e.g., pants/skirt type, color), \textbf{Accessories} (e.g., bag, hat, glasses), and \textbf{Extrinsic}, which merges Up, Down, and Accessories clusters. By systematically including or excluding these clusters and their constituent attributes, we generated a diverse set of experimental configurations. In addition, we included every single-attribute experiment, as well as some extra combinations. For each configuration, we computed two standard ReID metrics, namely mean Average Precision (mAP) and Rank-1 accuracy (R1).

\begin{figure*}[t]
    \centering
    \includegraphics[width=0.9\textwidth]{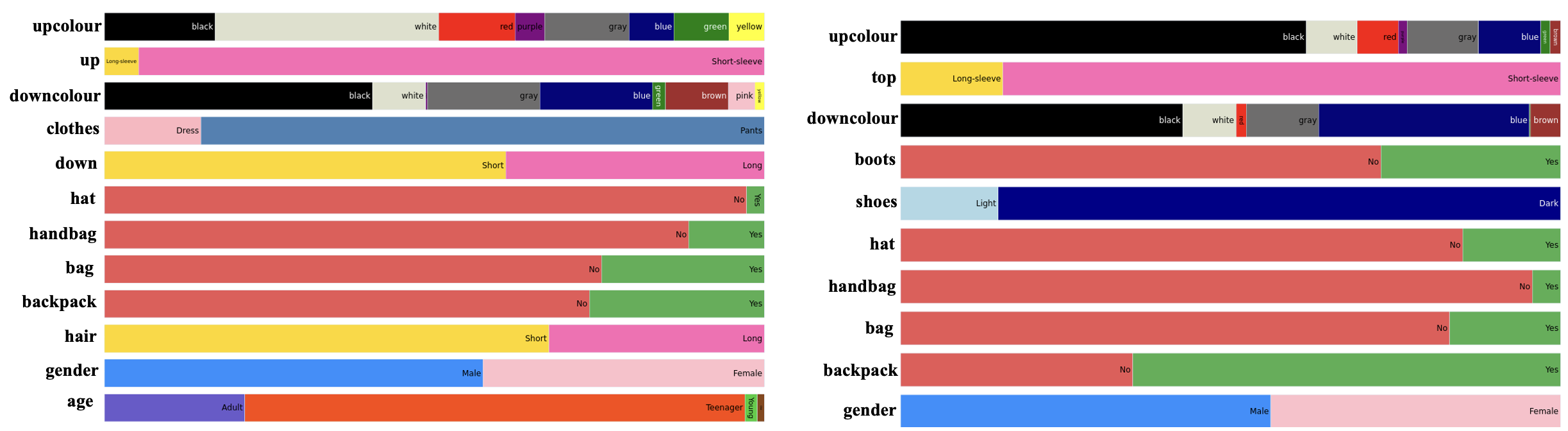}
    \caption{\textbf{Prior probabilities} for each value of the manually annotated attributes in the Market-1501 (\cite{market, market-attr}) (left) and DukeMTMC (\cite{dukemtmc, dukemtmc-attr}) (right) datasets. For each dataset, the horizontal bars represent the distribution of attribute values across all annotated images, with color segments indicating the proportion of each category (e.g., gender, clothing type, color, accessories). }
    \label{fig:attr-distribution}
\end{figure*}

Figure \ref{fig:attr-distribution} visualizes the prior probabilities for each value of the manually annotated attributes in both datasets. Notably, certain attributes display strong class imbalance. Variation in prevalence , such as specific colour dominance or specific accessories scarcity, highlights the importance of considering baseline attribute distributions when interpreting model performance and the relative impact of diverse semantic cues. In general, the skewed frequencies observed across intrinsic (e.g., gender, age), upper- and lower-body, and accessory groups indicate that simple dataset priors may influence the learned prominence of these attributes in downstream tasks. Thus, understanding and accounting for these distributional biases is essential for drawing fair, interpretable conclusions about attribute importance in person ReID. In the evaluation, these priors inform our interpretation of attribute importance, providing critical context for the statistical analyses.

\subsection{Implementation details}
\label{results_implementationdetails}

\noindent MoSAIC-ReID is trained using a Vision Transformer backbone (ViT-B-16) on a single 24Gb Nvidia RTX4090 GPU for 120 epochs. We opted for implementing our module using LoRA rank 16, \textit{K}=12 and mean pooling (see supplementary materials for ablation). Optimization uses AdamW (\cite{loshchilov2017decoupled}) with a base learning rate of 0.0001. Learning rate decays by a factor of 0.1 at epochs 90 and 110. 

\subsection{Person Re-Identification Performance}
\label{results_reidresults}

\input{tables/related-work}

\noindent We benchmark MoSAIC-ReID against both general and attribute-based ReID methods on Market-1501 and DukeMTMC (Table \ref{tab:related-work}). While our framework improves upon the CLIP-ReID baseline when attribute annotations are available, these gains \textit{should not be interpreted as directly comparable to approaches} that do not require attribute supervision at inference. Our \textbf{primary objective} is not to compete on raw ReID accuracy but to \textbf{provide a systematic framework for quantifying the relative importance of semantic attributes in driving identification performance}. Performance gains serve as a validation signal for the framework’s utility, rather than the central goal.
\input{tables/experiments}

We conducted experiments on both datasets(Table \ref{tab:experiments}) using a diverse set of attribute inclusion combinations to illustrate the relative importance of individual or group of attributes. Including rich semantic clusters such as \textit{upper} and \textit{lower} body clothing colors consistently yield the highest retrieval accuracy, while accessories provide notable boosts. Removing key attributes results in measurable degradation, confirming the utility of attribute-informed routing in disentangling and leveraging fine-grained cues.

\subsection{Attribute importance}
\label{results_attrimportance}

\noindent To quantify the influence of individual semantic attributes on person re-identification performance, we employed also three complementary methods: a General Linear Model (GLM) for statistically controlled effect estimation, a machine learning-based approach to capture non-linear ranking patterns and classical hypothesis testing to validate attribute significance. 
In our statistical analyses, \textbf{the unit of observation is the experiment configuration}, each of which yields a pair of evaluation metrics (mAP and Rank‑1). Thus, each configuration represents a single observation in the analyses, rather than individual query–gallery image pairs.

First, we applied GLM (\cite{nelder1972generalized}) to estimate the independent effect of each attribute. Each attribute was encoded as a binary variable reflecting its inclusion in the test-time configuration. GLM outputs interpretable coefficients that quantify the unique contribution of each attribute to mAP, while controlling for correlations among attributes. This method revealed strong positive contributions from attributes such as \textit{age}, \textit{upcolour}, and \textit{downcolour} in Market-1501, and \textit{gender}, \textit{downcolour}, and \textit{top} in DukeMTMC (Table \ref{tab:glm}). In both datasets, color-related cues for clothing demonstrated the highest effect sizes with strong statistical significance, while attributes like \textit{hat}, \textit{handbag}, and \textit{up} showed weak or non-significant contributions, with wide confidence intervals crossing zero, suggesting lower discriminative power or under-representation in the data. Notably, \textit{backpack} and \textit{bag} produced moderate positive effects in both datasets, reflecting their utility when present, despite being less frequent. These results support our broader hypothesis that both attribute salience and distributional properties influence their relevance for ReID tasks. Experiments of effects of each attribute across all rank-1 scores for both datasets yielded similar conclusions (supplementary material). 

\input{tables/glm}

\begin{figure*}[ht!]
    \centering
    \includegraphics[width=0.41\linewidth]{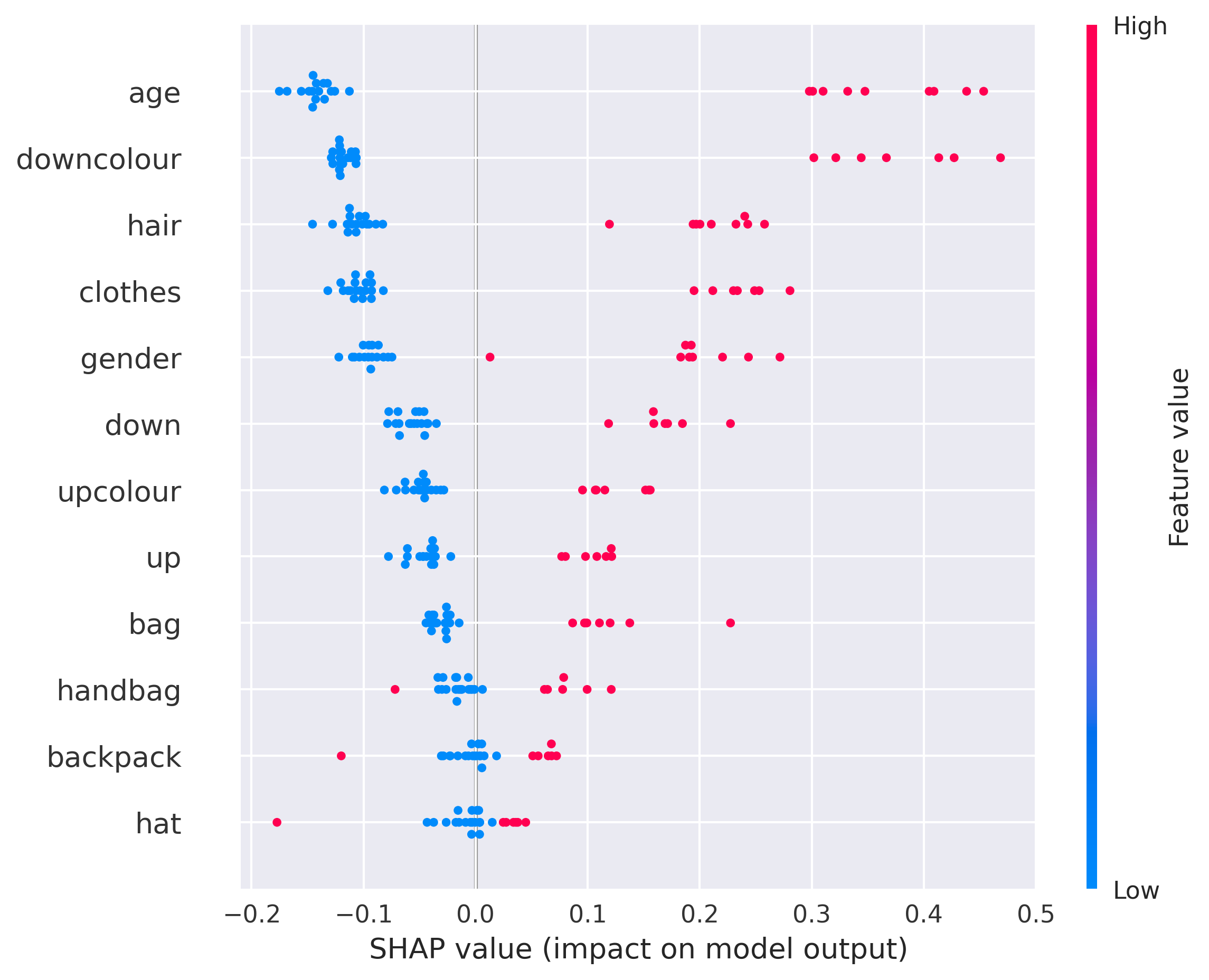}
    \includegraphics[width=0.48\linewidth]{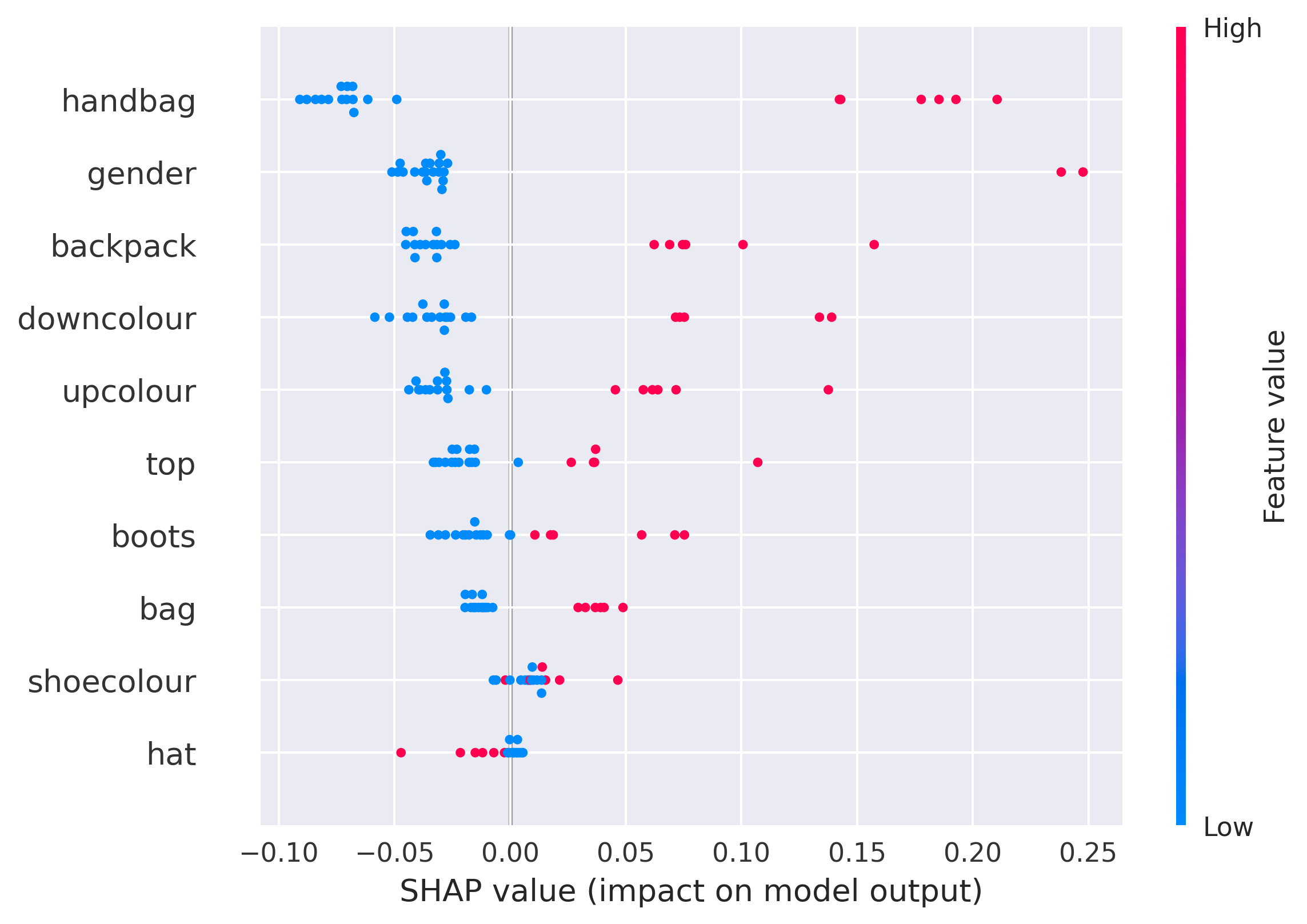}

    \caption{\textbf{SHAP scores for feature importance} on Market1501 (\cite{market}) (left) and DukeMTMC (\cite{dukemtmc}) (right) datasets.}
    \label{fig:supp-shap}
\end{figure*}

To complement the linear model, we trained a Random Forest (RF) regressor (\cite{ho1995random}) to predict mAP and rank-1 based on attribute inclusion patterns, since these models are capable of modeling non-linear relationships and provide intrinsic Feature Importance (FIMP) estimates. To enhance interpretability, we further applied Permutation Feature Importance (PIMP) (\cite{fisher2019all}) and SHapley Additive exPlanations (SHAP) values (\cite{lundberg2017unified}). PIMP quantified the effect of shuffling each attribute on prediction error, while SHAP scores provided local and global attribution. Both analyses consistently identified \textit{downcolour} as the key contributor in both Market-1501 and DukeMTMC datasets. In Market-1501, attributes like \textit{upcolour}, \textit{age} and \textit{bag} also showed substantial contribution, especially in SHAP scores, while in DukeMTMC, \textit{gender} and \textit{upcolour} had notable importance. Results are summarized in Figure \ref{fig:supp-shap} and Table \ref{tab:feat-importance}

\input{tables/feature-importance}

Lastly, to assess significance from a classical statistical perspective, we conducted independent two-sample t-tests comparing the mean mAP between experiments of inclusion versus exclusion for each attribute to identify attributes whose inclusion led to statistically significant changes in performance. Intrinsic and garments colour attributes in Market1501 and the presence of certain accessories in DukeMTMC showed highly significant effects ($p < 0.01$), reinforcing the findings from the model-based analyses. For both datasets, the most substantial t-statistics and lowest p-values were consistently observed for color-related cues and notable accessories, suggesting these features provide robust, discriminative signals for person re-identification. 

Overall, these findings consistently highlight the critical role of certain semantic attributes, particularly clothing colour and certain accessories, in driving performance across datasets. Among colour ones, lower clothing colour appears to exert a stronger effect, suggesting that lower-body appearance provides more distinctive cues for discrimination in these benchmarks. Accessory-related attributes, meanwhile, show variable importance closely tied to their prevalence within each dataset. For example, the \textit{backpack} attribute in DukeMTMC is relatively frequent and exhibits significant positive impact, whereas the \textit{hat} attribute (scarce in both) displays diminished importance, likely due to insufficient representation. Crucially, this multifaceted analysis is made possible by our framework, which modularly isolates attribute contributions through semantically aligned experts, enabling principled measurement of what attributes truly matter for person ReID. This offers not only deeper interpretability but a concrete guideline for integrating semantic priors during deployment in real-world applications. Although configurations share the same test set, randomization in training and model initialization, along with aggregation over multiple queries, mitigate dependency effects. Nonetheless, we acknowledge that some dependency may exist, and future work could explore statistical methods robust to correlated measurements.

%% file: tables/attributes.tex
\begin{table}[htbp]
\centering
\label{tab:attributes}
\resizebox{0.8\linewidth}{!}{%
\begin{tabular}{@{}l l l l c c@{}}
\toprule
Supercategory & Attribute& Description & Type & DukeMTMC & Market1501 \\
\midrule
\multirow{3}{*}{Intrinsic} & age & Age group & M & \ding{56} & \ding{52} \\
& gender & Gender & DS-B & \ding{52} & \ding{52} \\
& hair & Hair length & DS-B & \ding{56} & \ding{52} \\
\midrule
\multirow{4}{*}{Accessories} & backpack & Carrying backpack & SS-B & \ding{52} & \ding{52} \\
& bag & Carrying bag & SS-B & \ding{52} & \ding{52} \\
& handbag & Carrying handbag & SS-B & \ding{52} & \ding{52} \\
& hat & Wearing hat & SS-B & \ding{52} & \ding{52} \\
\midrule
\multirow{2}{*}{Upper-body} & top/up & Upper length/sleeve & DS-B & \ding{52} & \ding{52} \\
& upcolour & Upper color & M & \ding{52} & \ding{52} \\
\midrule
\multirow{5}{*}{Lower-body} & boots & Wearing boots & SS-B & \ding{52} & \ding{56} \\
& shoes & Shoe color & DS-B & \ding{52} & \ding{56} \\
& down & Lower length & DS-B & \ding{56} & \ding{52} \\
& clothes & Lower type & DS-B & \ding{56} & \ding{52} \\
& downcolour & Lower color & M & \ding{52} & \ding{52} \\
\bottomrule
\end{tabular}%
}

\caption{\textbf{Overview of semantic attributes in this study.} \textit{SS-B}: single-state binary, \textit{DS-B}: dual-state binary, \textit{M}: multiclass.}
\label{table-attr}
\end{table}

%% file: tables/related-work.tex
\begin{table}[ht!]
\centering
\resizebox{0.8\linewidth}{!}{
\begin{tabular}{l|cc|cc}
\toprule
\multirow{2}{*}{\textbf{Model}}  & \multicolumn{2}{c|}{\textbf{Market1501}} & \multicolumn{2}{c}{\textbf{DukeMTMC}} \\
  & mAP & Rank-1 & mAP & Rank-1 \\
\midrule
\multicolumn{5}{l}{\textit{Attribute-based methods}} \\
\hline
AttriVision \cite{sedeh2025attrivision} & 83.8 & 88.8 & ---     & ---   \\
MPV$^2$P \cite{dong2024multi}             & 87.9 & 95.3 & 80.9    & 91.9   \\
AGCL \cite{zhang2023attribute}          & 88.1 & 95.8 & ---     & ---   \\
ADR \cite{shi2022attribute}             & 88.5 & 95.7 & 79.0    & 89.3 \\
Cerberus \cite{eom2025cerberus}         & 89.8 & 96.1 & 80.7    & 91.1  \\
MoSCE-ReID \cite{ren2025mosce}          & 94.2 & 97.9 & 87.3    & 94.6  \\
\midrule
\multicolumn{5}{l}{\textit{General Methods}} \\
\hline
AAFormer \cite{zhu2023aaformer}         & 88.0 & 95.4 & 80.9    & 90.1  \\
CLIMB-ReID  \cite{yu2025climb}          & 92.6 & 96.8 & ---     & ---   \\
SOLIDER  \cite{chen2023beyond}          & 93.9 & 96.9 & ---     & ---   \\
PromptSG \cite{yang2024pedestrian}      & 94.6 & 97.0 & 81.6    & 91.0  \\
CLIP-ReID (baseline) \cite{li2023clip}  & 89.6 & 95.5 & 82.5    & 90.0  \\
\textbf{MoSAIC-ReID (Ours)}             & 95.5 & 97.9 & 85.7    & 93.3 \\
\bottomrule
\end{tabular}
}
\caption{\textbf{Re-identification results} (without re-ranking).}
\label{tab:related-work}
\end{table}

%% file: tables/experiments.tex
\begin{table}[htbp]
\centering
\resizebox{\linewidth}{!}{%
\begin{tabular}{@{}ccccccccccccccc@{}}
\toprule
Gender & Hair & Age & Hat & Backpack & Bag & Handbag & Up & Upcolour & Down & Downcolour & Clothes & mAP & rank-1 \\
\midrule
\ding{52} & \ding{52} & \ding{52} & \ding{52} & \ding{52} & \ding{52} & \ding{52} & \ding{52} & \ding{52} & \ding{52} & \ding{52} & \ding{52} & \textbf{95.5} & \textbf{97.9} \\
\ding{52} & \ding{52} & \ding{52} & \ding{56} & \ding{56} & \ding{56} & \ding{56} & \ding{56} & \ding{56} & \ding{56} & \ding{56} & \ding{56} & 90.8 & 96.2 \\
\ding{56} & \ding{56} & \ding{56} & \ding{52} & \ding{52} & \ding{52} & \ding{52} & \ding{52} & \ding{52} & \ding{52} & \ding{52} & \ding{52} & 94.6 & 97.6 \\
\ding{52} & \ding{52} & \ding{52} & \ding{56} & \ding{56} & \ding{56} & \ding{56} & \ding{52} & \ding{52} & \ding{56} & \ding{56} & \ding{56} & 91.9 & 96.7 \\
\ding{52} & \ding{52} & \ding{52} & \ding{56} & \ding{56} & \ding{56} & \ding{56} & \ding{56} & \ding{56} & \ding{52} & \ding{52} & \ding{52} & 92.7 & 97.5 \\
\ding{52} & \ding{52} & \ding{52} & \ding{52} & \ding{52} & \ding{52} & \ding{52} & \ding{56} & \ding{56} & \ding{56} & \ding{56} & \ding{56} & 92.4 & 96.6 \\
\ding{56} & \ding{56} & \ding{56} & \ding{56} & \ding{56} & \ding{56} & \ding{56} & \ding{56} & \ding{56} & \ding{52} & \ding{52} & \ding{52} & 91.3 & 96.4 \\
\ding{56} & \ding{56} & \ding{56} & \ding{52} & \ding{52} & \ding{52} & \ding{52} & \ding{56} & \ding{56} & \ding{56} & \ding{56} & \ding{56} & 91.2 & 95.8 \\
\ding{56} & \ding{56} & \ding{56} & \ding{56} & \ding{56} & \ding{56} & \ding{56} & \ding{52} & \ding{52} & \ding{56} & \ding{56} & \ding{56} & 90.7 & 95.6 \\
\ding{56} & \ding{56} & \ding{56} & \ding{56} & \ding{56} & \ding{56} & \ding{56} & \ding{52} & \ding{56} & \ding{56} & \ding{56} & \ding{56} & 89 & 95.2 \\
\ding{56} & \ding{56} & \ding{56} & \ding{56} & \ding{56} & \ding{56} & \ding{56} & \ding{56} & \ding{52} & \ding{56} & \ding{56} & \ding{56} & 90.3 & 95.3 \\
\ding{56} & \ding{56} & \ding{56} & \ding{56} & \ding{56} & \ding{56} & \ding{56} & \ding{56} & \ding{56} & \ding{56} & \ding{52} & \ding{56} & 90.6 & 95.6 \\
\ding{56} & \ding{56} & \ding{56} & \ding{56} & \ding{56} & \ding{56} & \ding{56} & \ding{56} & \ding{56} & \ding{52} & \ding{56} & \ding{56} & 88.5 & 95.0 \\
\ding{56} & \ding{56} & \ding{56} & \ding{56} & \ding{56} & \ding{56} & \ding{56} & \ding{56} & \ding{56} & \ding{56} & \ding{56} & \ding{52} & 88.9 & 95.1 \\
\ding{56} & \ding{56} & \ding{56} & \ding{56} & \ding{56} & \ding{52} & \ding{56} & \ding{56} & \ding{56} & \ding{56} & \ding{56} & \ding{56} & 89.6 & 95.2 \\
\ding{56} & \ding{56} & \ding{56} & \ding{56} & \ding{56} & \ding{56} & \ding{52} & \ding{56} & \ding{56} & \ding{56} & \ding{56} & \ding{56} & 89.2 & 94.8 \\
\ding{56} & \ding{56} & \ding{56} & \ding{52} & \ding{56} & \ding{56} & \ding{56} & \ding{56} & \ding{56} & \ding{56} & \ding{56} & \ding{56} & 88.6 & 94.6 \\
\ding{56} & \ding{56} & \ding{56} & \ding{56} & \ding{52} & \ding{56} & \ding{56} & \ding{56} & \ding{56} & \ding{56} & \ding{56} & \ding{56} & 89.4 & 94.8 \\
\ding{56} & \ding{52} & \ding{56} & \ding{56} & \ding{56} & \ding{56} & \ding{56} & \ding{56} & \ding{56} & \ding{56} & \ding{56} & \ding{56} & 89 & 95.1 \\
\ding{56} & \ding{56} & \ding{52} & \ding{56} & \ding{56} & \ding{56} & \ding{56} & \ding{56} & \ding{56} & \ding{56} & \ding{56} & \ding{56} & 89.8 & 95.4 \\
\ding{52} & \ding{56} & \ding{56} & \ding{56} & \ding{56} & \ding{56} & \ding{56} & \ding{56} & \ding{56} & \ding{56} & \ding{56} & \ding{56} & 88.5 & 94.6 \\
\ding{52} & \ding{52} & \ding{52} & \ding{56} & \ding{56} & \ding{56} & \ding{56} & \ding{52} & \ding{52} & \ding{52} & \ding{52} & \ding{52} & 93.8 & 97.6 \\
\ding{52} & \ding{52} & \ding{52} & \ding{52} & \ding{52} & \ding{52} & \ding{52} & \ding{52} & \ding{52} & \ding{56} & \ding{56} & \ding{56} & 93.6 & 97.4 \\
\ding{52} & \ding{52} & \ding{52} & \ding{52} & \ding{52} & \ding{52} & \ding{52} & \ding{56} & \ding{56} & \ding{52} & \ding{52} & \ding{52} & 94.5 & 97.5 \\
\ding{56} & \ding{56} & \ding{56} & \ding{56} & \ding{56} & \ding{56} & \ding{56} & \ding{56} & \ding{56} & \ding{56} & \ding{56} & \ding{56} & \underline{89.6} & \underline{95.5} \\
\bottomrule
\end{tabular}%
}

\vspace{1ex} 

\resizebox{\linewidth}{!}{%
\begin{tabular}{@{}cccccccccccc@{}}
\toprule
Backpack & Bag & Handbag & Hat & Boots & Shoe Color & Top & Gender & Down Color & Up Color & mAP & rank-1 \\
\midrule
\ding{52} & \ding{52} & \ding{52} & \ding{52} & \ding{52} & \ding{52} & \ding{52} & \ding{52} & \ding{52} & \ding{52} & \textbf{85.7} & \textbf{93.3} \\
\ding{52} & \ding{52} & \ding{52} & \ding{52} & \ding{52} & \ding{52} & \ding{56} & \ding{56} & \ding{56} & \ding{56} & 83.7 & 92.0 \\
\ding{56} & \ding{56} & \ding{56} & \ding{56} & \ding{56} & \ding{56} & \ding{52} & \ding{56} & \ding{56} & \ding{52} & 83.1 & 91.9 \\
\ding{56} & \ding{56} & \ding{56} & \ding{56} & \ding{52} & \ding{52} & \ding{56} & \ding{56} & \ding{52} & \ding{56} & 83.7 & 92.2 \\
\ding{56} & \ding{56} & \ding{56} & \ding{56} & \ding{56} & \ding{56} & \ding{56} & \ding{52} & \ding{56} & \ding{56} & 82.6 & 91.8 \\
\ding{56} & \ding{56} & \ding{56} & \ding{56} & \ding{56} & \ding{56} & \ding{52} & \ding{56} & \ding{56} & \ding{56} & 82.6 & 92.1 \\
\ding{56} & \ding{56} & \ding{56} & \ding{56} & \ding{56} & \ding{56} & \ding{56} & \ding{56} & \ding{56} & \ding{52} & 82.9 & 92.1 \\
\ding{56} & \ding{56} & \ding{56} & \ding{56} & \ding{56} & \ding{56} & \ding{56} & \ding{56} & \ding{52} & \ding{56} & 83.0 & 91.9 \\
\ding{56} & \ding{56} & \ding{56} & \ding{52} & \ding{56} & \ding{56} & \ding{56} & \ding{56} & \ding{56} & \ding{56} & 82.5 & 91.6 \\
\ding{52} & \ding{56} & \ding{56} & \ding{56} & \ding{56} & \ding{56} & \ding{56} & \ding{56} & \ding{56} & \ding{56} & 82.8 & 92.1 \\
\ding{56} & \ding{52} & \ding{56} & \ding{56} & \ding{56} & \ding{56} & \ding{56} & \ding{56} & \ding{56} & \ding{56} & 82.6 & 91.8 \\
\ding{56} & \ding{56} & \ding{52} & \ding{56} & \ding{56} & \ding{56} & \ding{56} & \ding{56} & \ding{56} & \ding{56} & 82.6 & 92.2 \\
\ding{56} & \ding{56} & \ding{56} & \ding{56} & \ding{56} & \ding{56} & \ding{52} & \ding{56} & \ding{52} & \ding{52} & 83.5 & 92.0 \\
\ding{52} & \ding{52} & \ding{52} & \ding{52} & \ding{56} & \ding{56} & \ding{52} & \ding{56} & \ding{56} & \ding{52} & 83.6 & 92.3 \\
\ding{52} & \ding{52} & \ding{52} & \ding{52} & \ding{52} & \ding{52} & \ding{56} & \ding{56} & \ding{52} & \ding{52} & 84.3 & 92.3 \\
\ding{56} & \ding{56} & \ding{56} & \ding{56} & \ding{52} & \ding{52} & \ding{56} & \ding{56} & \ding{56} & \ding{56} & 83.0 & 91.8 \\
\ding{52} & \ding{52} & \ding{52} & \ding{52} & \ding{56} & \ding{56} & \ding{56} & \ding{56} & \ding{56} & \ding{56} & 83.4 & 92.3 \\
\ding{56} & \ding{56} & \ding{56} & \ding{56} & \ding{56} & \ding{52} & \ding{56} & \ding{56} & \ding{56} & \ding{56} & 82.5 & 91.6 \\
\ding{56} & \ding{56} & \ding{56} & \ding{56} & \ding{52} & \ding{56} & \ding{56} & \ding{56} & \ding{56} & \ding{56} & 82.6 & 91.9 \\
\ding{56} & \ding{56} & \ding{56} & \ding{56} & \ding{56} & \ding{56} & \ding{56} & \ding{56} & \ding{56} & \ding{56} & \underline{82.6} & \underline{91.6} \\
\bottomrule
\end{tabular}%
}
\caption{\textbf{Different attribute inclusion combinations results} for Market1501 (up) and DukeMTMC (down).}
\label{tab:experiments}
\end{table}

%% file: tables/glm.tex
\begin{table}
\centering
\resizebox{0.8\linewidth}{!}{%
\begin{tabular}{lrrrrrr}
 & Coef. & Std.Err. & z & P$>$$|$z$|$ & [0.025 & 0.975] \\
\midrule
\multicolumn{7}{c}{\textbf{Market 1501}} \\
\hline
Intercept & 88.881 & 0.114 & 773.412 & 0.000 & 88.656 & 89.107 \\
gender & -0.111 & 0.331 & -0.336 & 0.736 & -0.761 & 0.538 \\
hair & 0.388 & 0.331 & 1.170 & 0.241 & -0.261 & 1.038 \\
age & 1.188 & 0.331 & 3.581 & 0.000 & 0.538 & 1.838 \\
hat & -0.111 & 0.348 & -0.319 & 0.749 & -0.795 & 0.572 \\
backpack & 0.688 & 0.348 & 1.974 & 0.048 & 0.004 & 1.372 \\
bag & 0.888 & 0.348 & 2.547 & 0.010 & 0.204 & 1.572 \\
handbag & 0.488 & 0.348 & 1.400 & 0.161 & -0.195 & 1.172 \\
up & 0.078 & 0.298 & 0.263 & 0.792 & -0.507 & 0.664 \\
upcolour & 1.378 & 0.298 & 4.611 & 0.000 & 0.792 & 1.964 \\
down & -0.124 & 0.332 & -0.375 & 0.707 & -0.776 & 0.526 \\
downcolour & 1.975 & 0.332 & 5.943 & 0.000 & 1.323 & 2.626 \\
clothes & 0.275 & 0.332 & 0.828 & 0.407 & -0.376 & 0.926 \\
\hline
\multicolumn{7}{c}{\textbf{DukeMTMC}} \\
\hline
Intercept & 82.405 & 0.078 & 1046.590 & 0.000 & 82.251 & 82.559 \\
backpack & 0.369 & 0.211 & 1.745 & 0.080 & -0.045 & 0.784 \\
bag & 0.169 & 0.211 & 0.801 & 0.422 & -0.245 & 0.584 \\
handbag & 0.169 & 0.211 & 0.801 & 0.422 & -0.245 & 0.584 \\
hat & 0.069 & 0.211 & 0.329 & 0.741 & -0.345 & 0.484 \\
boots & 0.323 & 0.186 & 1.734 & 0.082 & -0.042 & 0.688 \\
shoecolour & 0.223 & 0.186 & 1.197 & 0.231 & -0.142 & 0.588 \\
top & 0.306 & 0.166 & 1.841 & 0.065 & -0.019 & 0.632 \\
gender & 0.483 & 0.187 & 2.573 & 0.010 & 0.115 & 0.851 \\
downcolour & 0.604 & 0.150 & 4.027 & 0.000 & 0.310 & 0.898 \\
upcolour & 0.285 & 0.167 & 1.705 & 0.088 & -0.042 & 0.613 \\
\bottomrule
\end{tabular}%
}
\caption{ \textbf{GLM regression results relative to mAP.}}
\label{tab:glm}
\end{table}

%% file: tables/feature-importance.tex
\begin{table}
\centering
\resizebox{0.7\linewidth}{!}{%
\begin{tabular}{lcrrll}
 & \multicolumn{2}{c}{\textbf{ML Analysis}} & \multicolumn{3}{c}{\textbf{Hypothesis testing}} \\
 & FIMP & PIMP & t-stat & p-value & Cohen's d \\
\midrule
\multicolumn{6}{c}{\textbf{Market 1501}} \\
\hline
gender      & 0.087 & 0.033 & -3.298 & 0.006**  & 1.509 \\
hair        & 0.065 & 0.023 & -3.549 & 0.003**  & 1.590 \\
age         & 0.143 & 0.087 & -3.981 & 0.001**  & 1.731 \\
hat         & 0.072 & 0.029 & -2.802 & 0.024*   & 1.542 \\
backpack    & 0.059 & 0.033 & -3.226 & 0.011*   & 1.686 \\
bag         & 0.079 & 0.043 & -3.341 & 0.009**  & 1.724 \\
handbag     & 0.065 & 0.025 & -3.115 & 0.014*   & 1.648 \\
up          & 0.065 & 0.034 & -2.578 & 0.031*   & 1.334 \\
upcolour    & 0.061 & 0.080 & -3.206 & 0.010**  & 1.542 \\
down        & 0.077 & 0.024 & -2.900 & 0.021*   & 1.630 \\
downcolour  & 0.141 & 0.158 & -4.203 & 0.002**  & 2.073 \\
clothes     & 0.086 & 0.046 & -3.114 & 0.015*   & 1.705 \\
\hline
\multicolumn{6}{c}{\textbf{DukeMTMC}} \\
\hline
backpack    & 0.085 & 0.075 & -2.555 & 0.046*   & 1.739  \\
bag         & 0.094 & 0.062 & -2.340 & 0.061    & 1.613  \\
handbag     & 0.092 & 0.062 & -2.340 & 0.061    & 1.613  \\
hat         & 0.061 & 0.028 & -2.237 & 0.071    & 1.554  \\
boots       & 0.084 & 0.054 & -2.088 & 0.086    & 1.443  \\
shoecolour  & 0.101 & 0.070 & -1.998 & 0.097    & 1.391  \\
top         & 0.055 & 0.047 & -1.300 & 0.255    & 0.967  \\
gender      & 0.113 & 0.095 & -0.704 & 0.608    & 1.503  \\
downcolour  & 0.247 & 0.256 & -2.451 & 0.064    & 1.926  \\
upcolour    & 0.067 & 0.090 & -2.251 & 0.067    & 1.498  \\
\bottomrule
\end{tabular}%
}
\caption{\textbf{Feature importance results.} */**: p-value $<$ 0.05/0.01}
\label{tab:feat-importance}
\end{table}

%% file: sec/conclusions.tex
\section{Conclusions}
\label{conclusions}

In this work, we introduce a modular MoE framework within a Transformer architecture to disentangle the impact of individual semantic attributes on person re-identification. LoRA experts are specialized per attribute, with an oracle-based router enabling controlled analysis of attribute importance. While our method shows competitive performance on Market-1501 and DukeMTMC when attribute annotations are available, its main contribution lies in systematically quantifying which attributes, primarily clothing colors and intrinsic characteristics, drive robust retrieval. Our results demonstrate that inference-time knowledge of key attributes can enhance ReID accuracy and robustness, suggesting that real-time systems equipped with external attribute predictors could benefit significantly. These findings are specific to our CLIP-ReID + MoSAIC-ReID framework; however, MoSAIC-ReID is modular and could yield comparable insights when integrated with other transformer-based ReID architectures. Overall, the framework highlights both the interpretability gains and practical requirements for integrating explicit semantic knowledge in real-world ReID applications.

%% file: sec/appendix.tex
\appendix
\section{Supplementary Material}
\label{app1}

Due to the limited space of the main paper, we provide
more experimental results and implementation details in the
supplementary material, which are additional figures for the results of the main paper, as well as, detailed results for the attribute analysis relative to the rank-1 metric.

\noindent\textbf{Last K ablation study.}Table~\ref{tab:lastk-results} reports performance on Market-1501\cite{market} when enabling MoSAIC-ReID modules in the last K transformer layers of ViT with all attributes considered. We observe consistent improvements in both mAP and rank-1 accuracy as K increases, with performance saturating around K=8. In our main experiments, we set K=12 to fully leverage expert integration, achieving the best overall results.

\input{tables/ablation-lk}

\noindent\textbf{Rank-1 attribute importance analysis. }Table \ref{tab:supp-glm-r1} refers to the GLM regression experiments to quantify the effects of each attribute across all rank-1 scores for Market1501 \cite{market} and DukeMTMC \cite{dukemtmc} datasets, where yielded similar conclusions were yielded in comparison to GLM experiments relative to mAP performance of MoSAIC-ReID. Again, the colour of lower garments and some intrinsic attributes such as \textit{age}, show significant positive effect on rank-1 performance.

\input{tables/supp/glm-rank1}

\begin{figure*}[h!]
    \centering
    \includegraphics[width=0.45\linewidth]{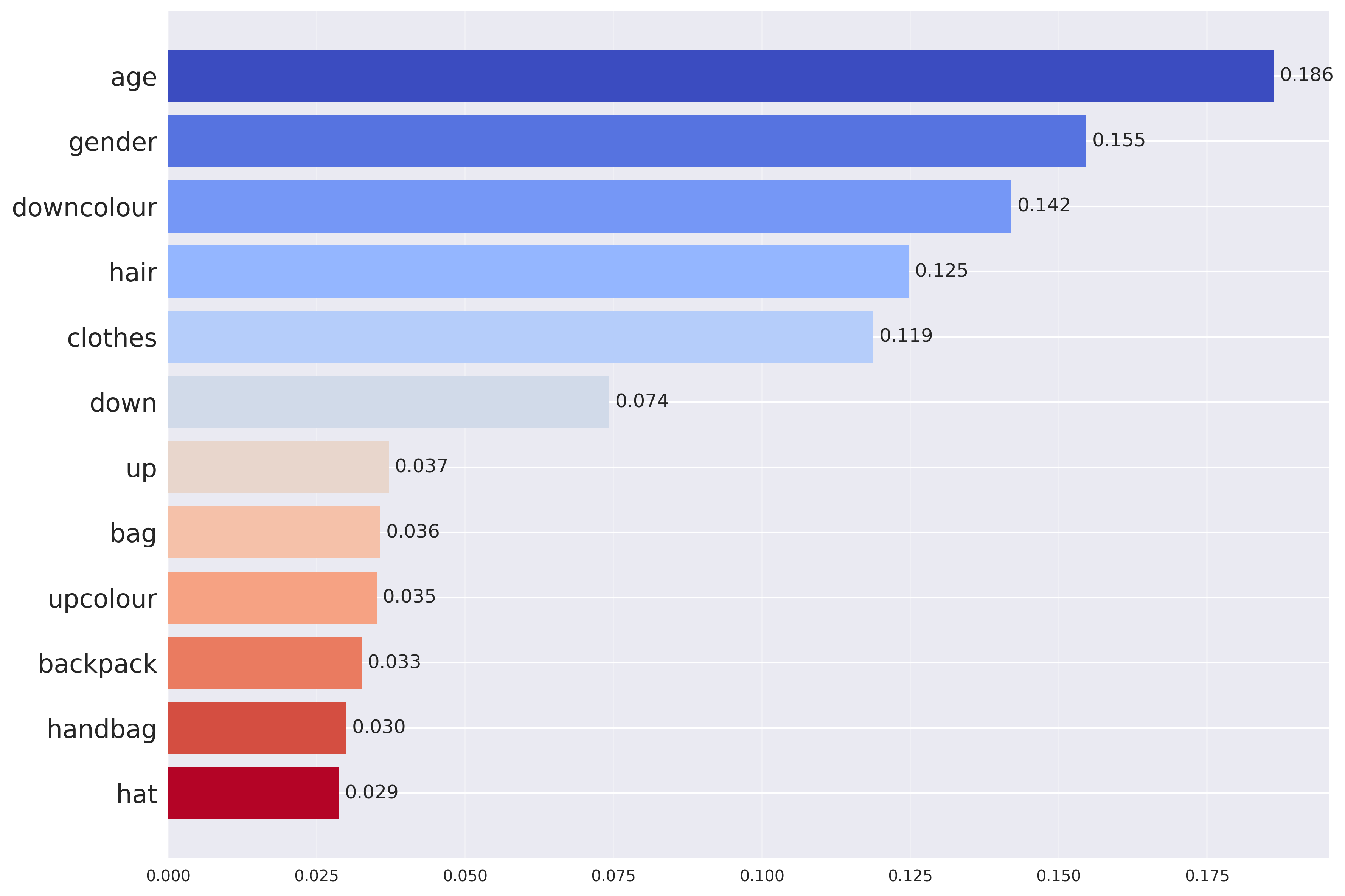}
    \includegraphics[width=0.45\linewidth]{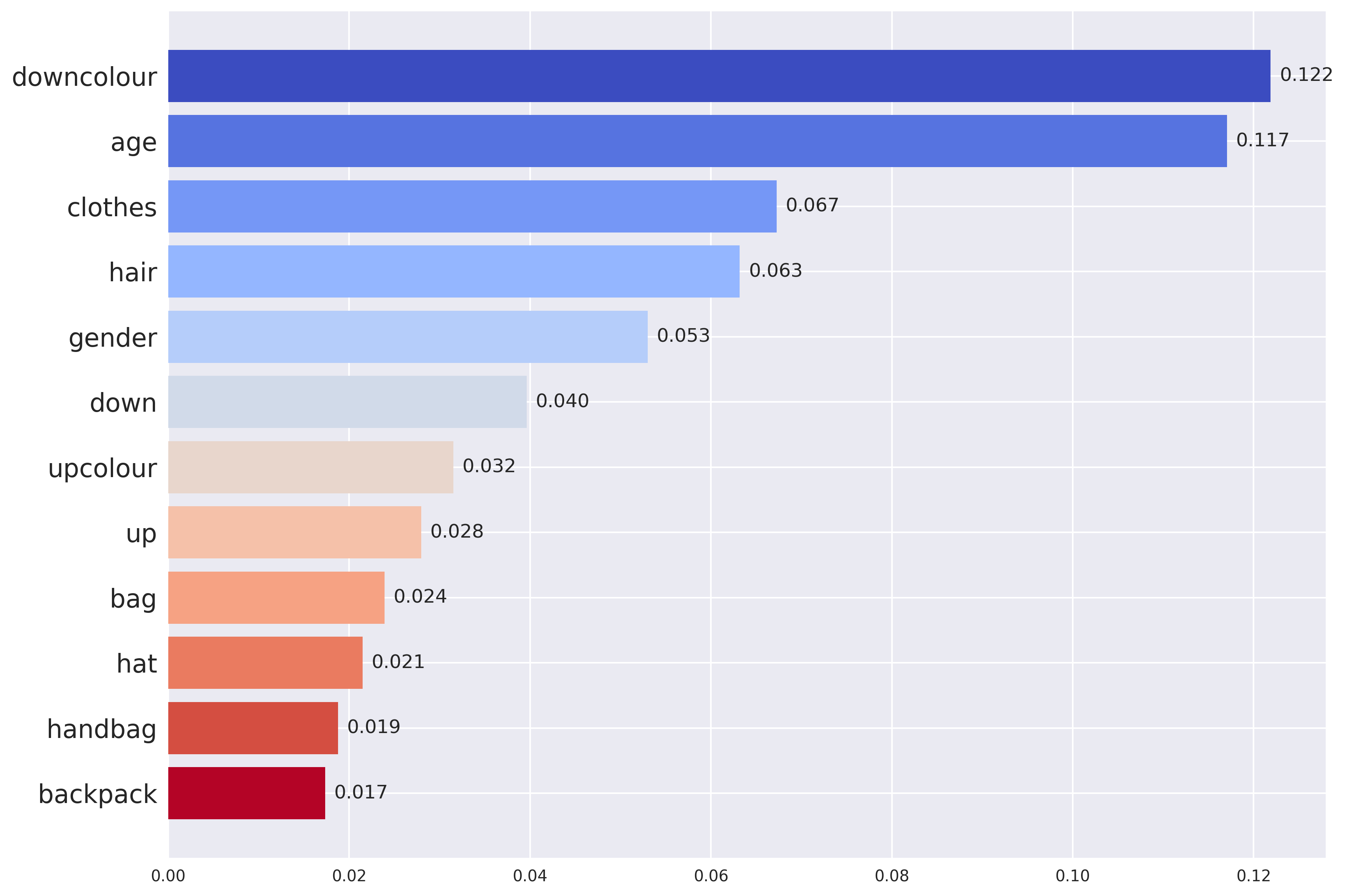}

    \caption{\textbf{RF Feature Importance (FIMP) \& Permutation Importance (PIMP) analysis} on Market1501\cite{market} dataset}
    \label{fig:supp-market-fimp}
\end{figure*}

\begin{figure*}[h!]
    \centering
    \includegraphics[width=0.45\linewidth]{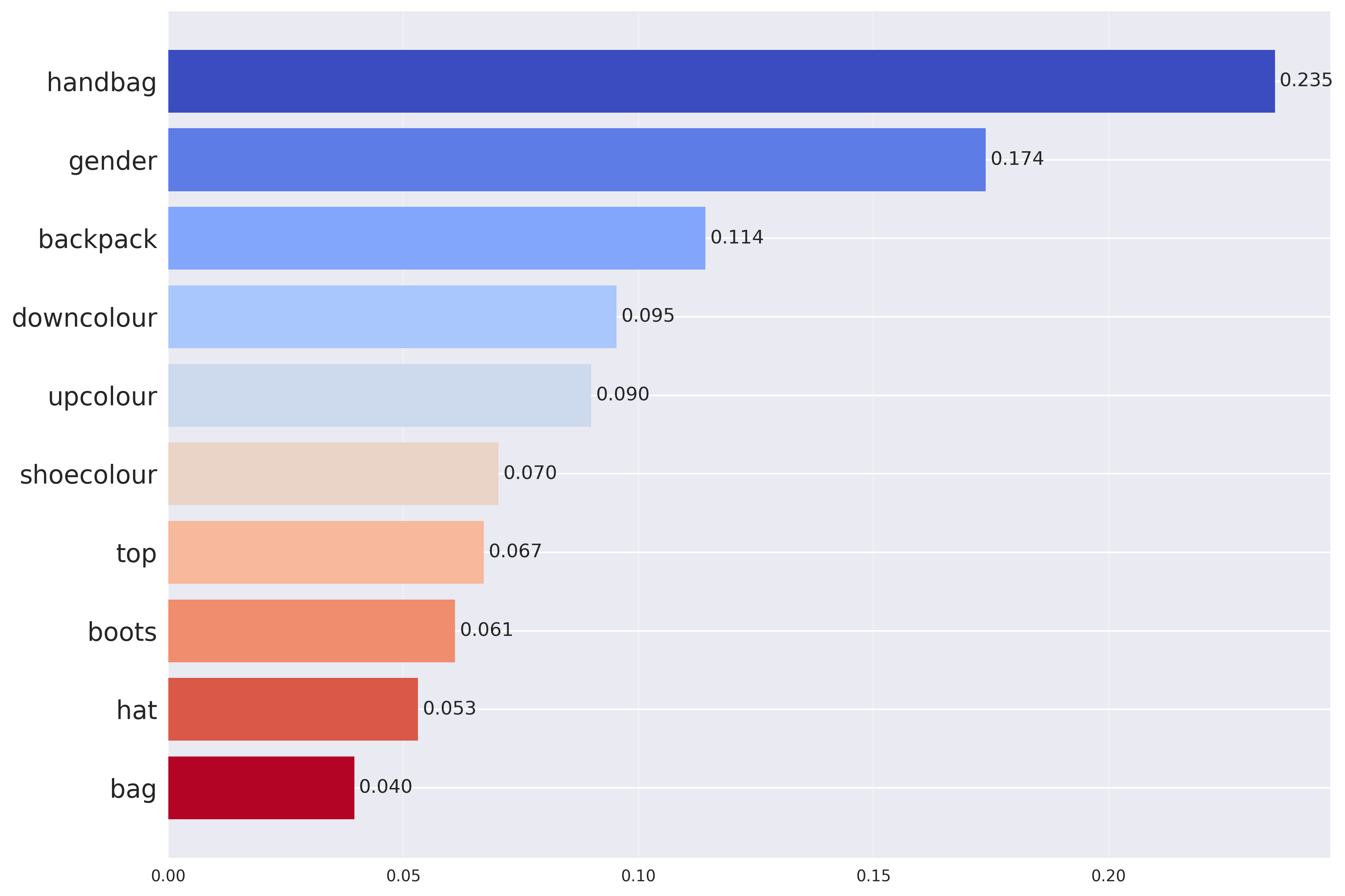}
    \includegraphics[width=0.45\linewidth]{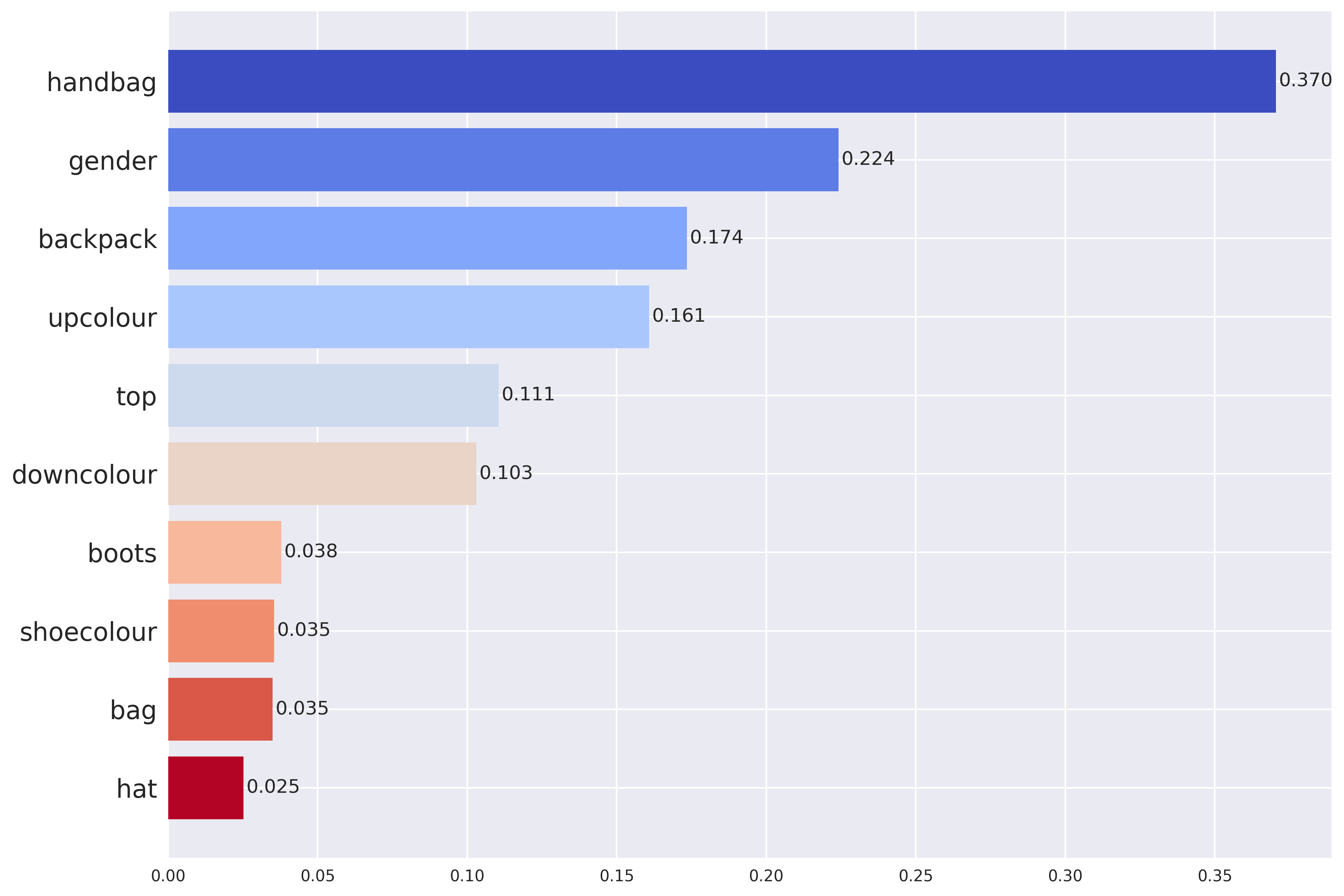}

    \caption{\textbf{RF Feature Importance (FIMP) \& Permutation Importance (PIMP) analysis} on DukeMTMC\cite{dukemtmc} dataset}
    \label{fig:supp-duke-fimp}
\end{figure*}

\begin{figure*}[h!]
    \centering
    \includegraphics[width=0.45\linewidth]{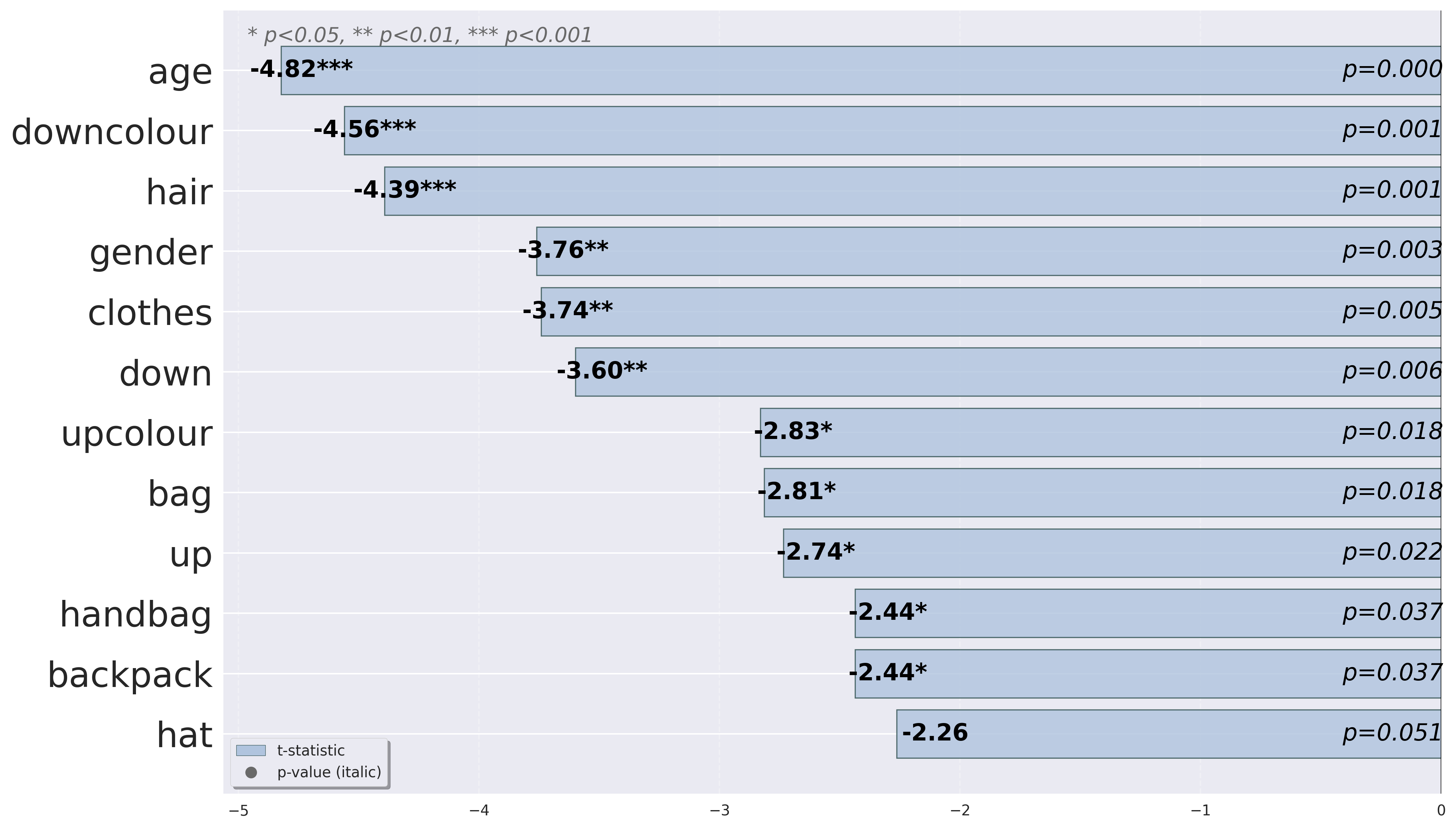}
    \includegraphics[width=0.45\linewidth]{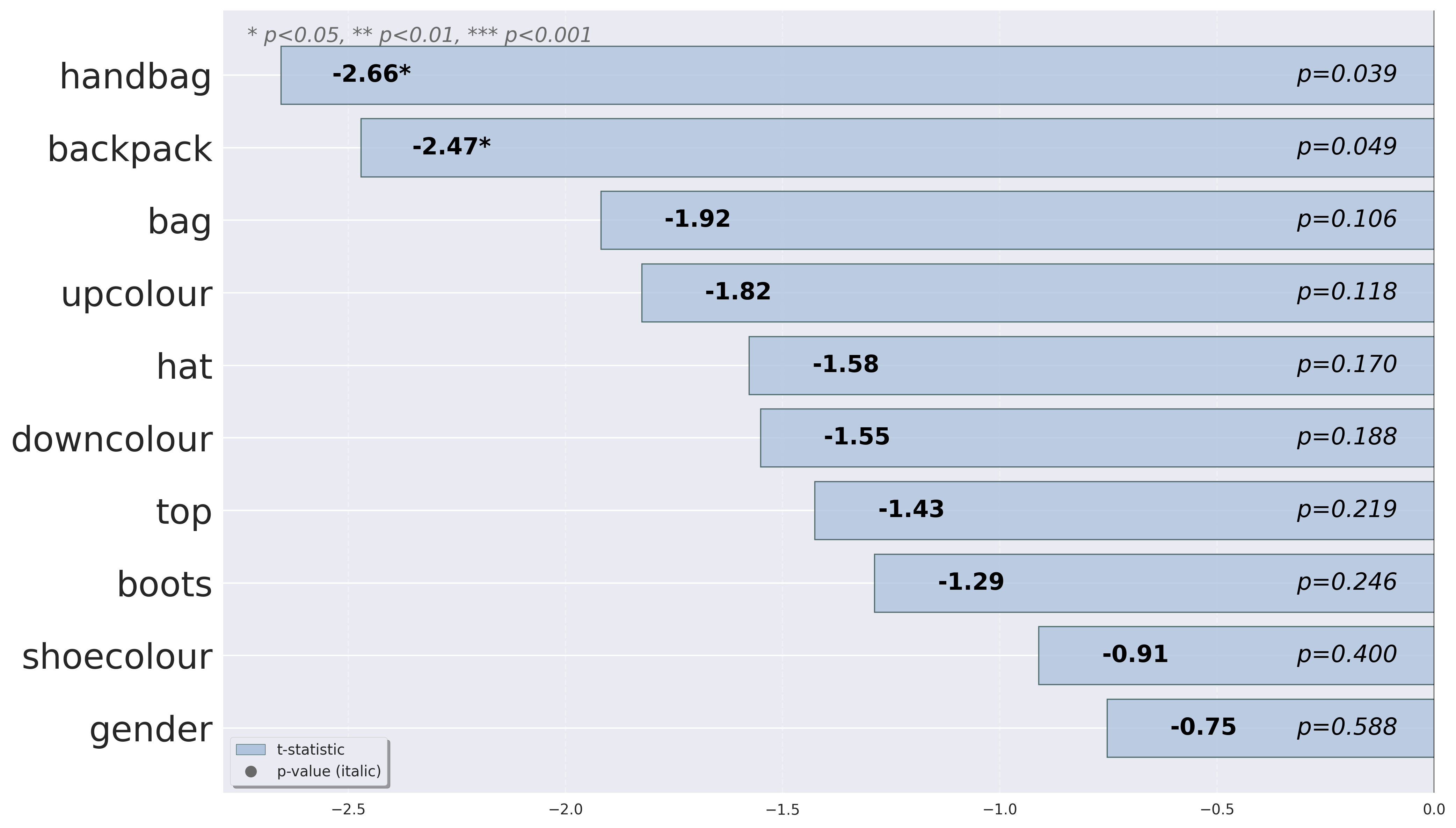}

    \caption{\textbf{Hypothesis testing (t-tests) for feature importance} on Market1501\cite{market} (left) and DukeMTMC\cite{dukemtmc} (right) datasets.}
    \label{fig:supp-ttest}
\end{figure*}

\begin{figure*}[h!]
    \centering
    \includegraphics[width=0.45\linewidth]{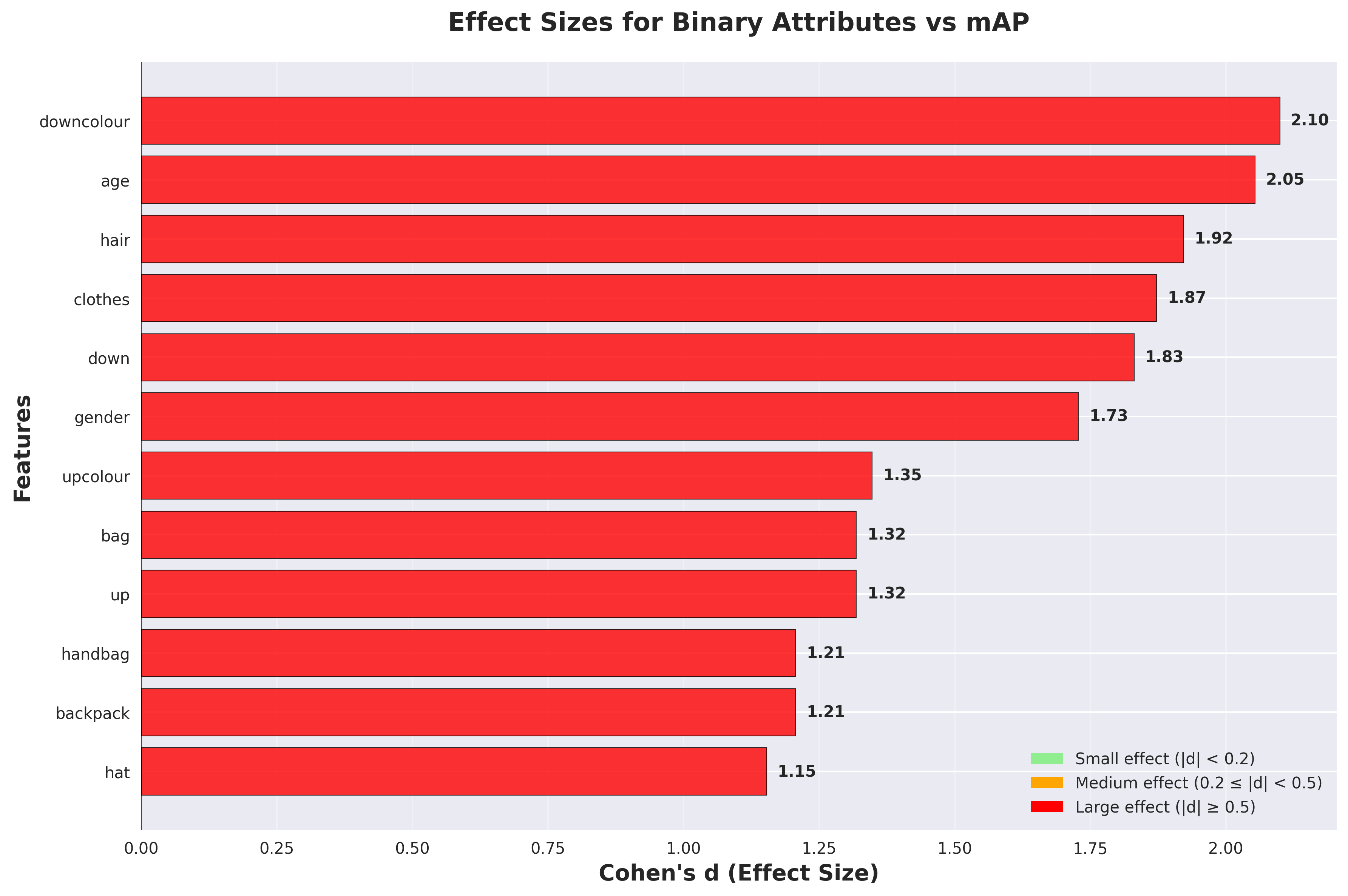}
    \includegraphics[width=0.45\linewidth]{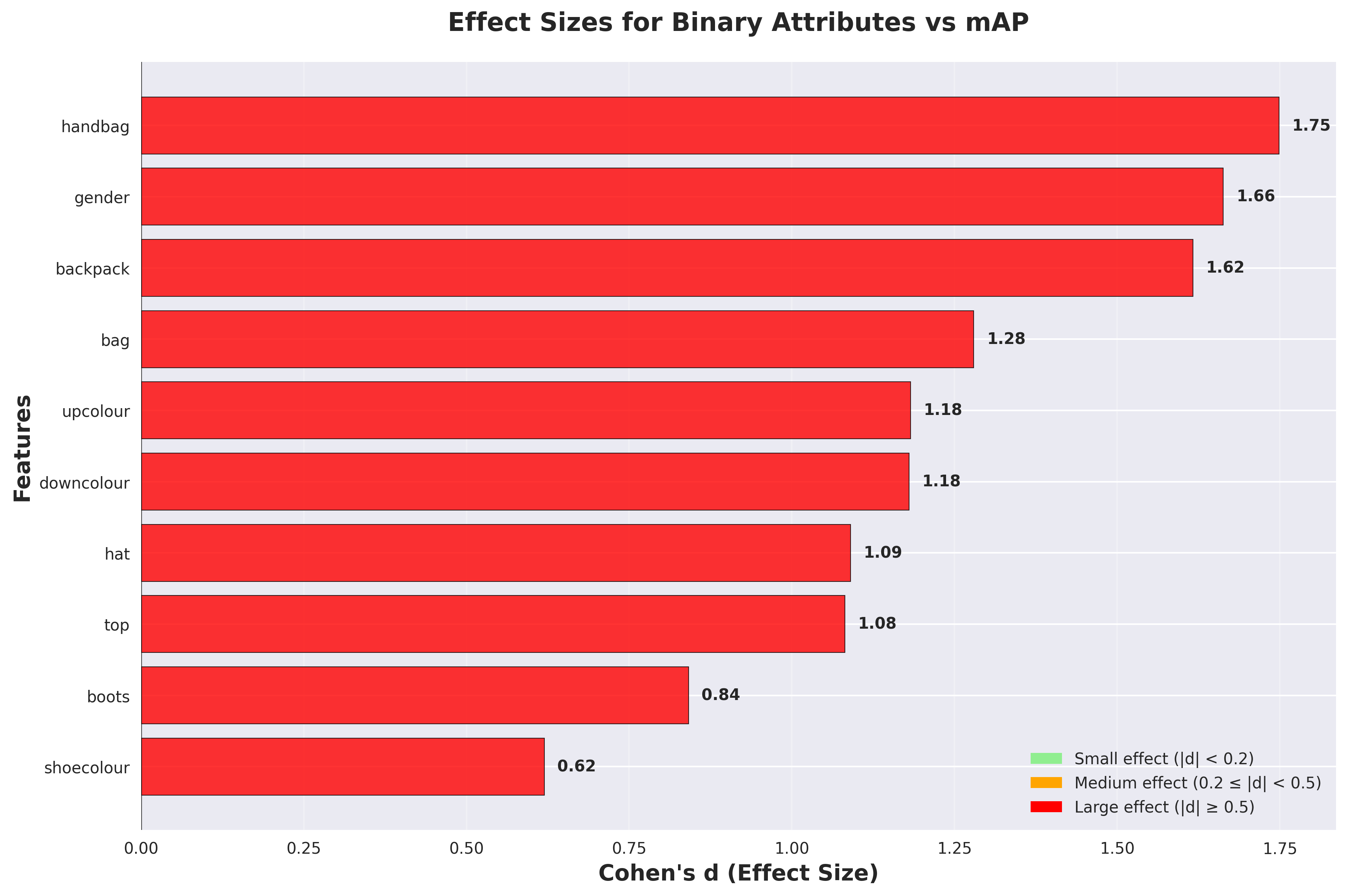}

    \caption{\textbf{Effect sizes (Cohen's d) from Hypothesis Testing} on Market1501\cite{market} (left) and DukeMTMC\cite{dukemtmc} (right) datasets.}
    \label{fig:supp-effect}
\end{figure*}

Figures~\ref{fig:supp-market-fimp}, \ref{fig:supp-duke-fimp}, \ref{fig:supp-shap} and \ref{fig:supp-ttest} present the RF feature importance rankings and hypothesis testing results for Rank-1 accuracy on Market1501 and DukeMTMC, respectively, all of which yield conclusions consistent with our mAP-based analysis—highlighting the strong and consistent impact of color-related attributes and certain accessories. Figure~\ref{fig:supp-effect} further quantifies these differences by reporting Cohen’s $d$ effect sizes, revealing that attributes like \textit{downcolour}, \textit{upcolour}, and \textit{age} have the largest standardized effects on rank-1 performance, thus reinforcing their discriminative value for person re-identification even under strict statistical criteria. 

\FloatBarrier

%% file: tables/ablation-lk.tex
\begin{table}[ht!]
\centering
\begin{tabular}{|c|c|c|}
\hline
\# last K & mAP & rank-1 \\
\hline
1     & 92.7 & 96.7 \\
2     & 93.7 & 97.2 \\
4     & 94.4 & 97.6 \\
8     & 95.3 & 98.0 \\
12    & 95.5 & 97.9 \\
\hline
\end{tabular}
\caption{Results on Market1501 for \textit{all} attributes for different number of last K layers where MoSAIC-ReID module is enabled.}
\label{tab:lastk-results}
\end{table}

%% file: tables/supp/glm-rank1.tex
\begin{table}
\centering
\resizebox{0.7\linewidth}{!}{%
\begin{tabular}{lrrrrrr}
\toprule
 & Coef. & Std.Err. & z & P$>$$|$z$|$ & [0.025 & 0.975] \\
\midrule
\multicolumn{7}{c}{\textbf{Market 1501}} \\
\hline
Intercept & 94.912 & 0.102 & 929.850 & 0.000 & 94.712 & 95.112 \\
gender & -0.093 & 0.294 & -0.316 & 0.751 & -0.670 & 0.484 \\
hair & 0.406 & 0.294 & 1.380 & 0.167 & -0.170 & 0.984 \\
age & 0.706 & 0.294 & 2.398 & 0.016 & 0.129 & 1.284 \\
hat & -0.104 & 0.309 & -0.337 & 0.736 & -0.711 & 0.502 \\
backpack & 0.095 & 0.309 & 0.308 & 0.757 & -0.511 & 0.702 \\
bag & 0.495 & 0.309 & 1.599 & 0.109 & -0.111 & 1.102 \\
handbag & 0.095 & 0.309 & 0.308 & 0.757 & -0.511 & 0.702 \\
up & 0.277 & 0.265 & 1.045 & 0.296 & -0.242 & 0.797 \\
upcolour & 0.377 & 0.265 & 1.421 & 0.155 & -0.142 & 0.897 \\
down & 0.164 & 0.295 & 0.556 & 0.577 & -0.414 & 0.742 \\
downcolour & 0.764 & 0.295 & 2.589 & 0.009 & 0.185 & 1.342 \\
clothes & 0.264 & 0.295 & 0.895 & 0.370 & -0.314 & 0.842 \\
\hline
\multicolumn{7}{c}{\textbf{DukeMTMC}} \\
\hline
Intercept & 91.743 & 0.078 & 1165.493 & 0.000 & 91.589 & 91.898 \\
backpack & 0.276 & 0.211 & 1.306 & 0.191 & -0.138 & 0.691 \\
bag & -0.023 & 0.211 & -0.110 & 0.912 & -0.438 & 0.391 \\
handbag & 0.376 & 0.211 & 1.778 & 0.075 & -0.038 & 0.791 \\
hat & -0.223 & 0.211 & -1.054 & 0.291 & -0.638 & 0.191 \\
boots & 0.186 & 0.186 & 0.999 & 0.317 & -0.178 & 0.551 \\
shoecolour & -0.113 & 0.186 & -0.610 & 0.541 & -0.478 & 0.251 \\
top & 0.215 & 0.166 & 1.295 & 0.195 & -0.110 & 0.541 \\
gender & 0.332 & 0.187 & 1.769 & 0.076 & -0.035 & 0.700 \\
downcolour & 0.213 & 0.149 & 1.424 & 0.154 & -0.080 & 0.507 \\
upcolour & 0.039 & 0.167 & 0.235 & 0.814 & -0.288 & 0.367 \\
\bottomrule
\end{tabular}%
}
\caption{ \textbf{GLM regression results vs Rank-1 metric.}}
\label{tab:supp-glm-r1}
\end{table}